\documentclass[
11pt, 
english, 
singlespacing, 
nolistspacing, 
liststotoc, 
headsepline, 
consistentlayout, 
openany,
]{MastersDoctoralThesis} 

\usepackage[utf8]{inputenc} 
\usepackage[T1]{fontenc} 

\usepackage{mathpazo} 
\usepackage{graphicx} 

\usepackage{makecell}
\usepackage{float}
\restylefloat{table}
\usepackage{hhline}
\usepackage{multirow}
\usepackage{colortbl}
\usepackage{tabularx}
\usepackage{booktabs}
\usepackage{hyperref}
\usepackage{amsmath}
\usepackage{amssymb}

\usepackage[backend=bibtex,style=authoryear,natbib=true]{biblatex} 

\usepackage[autostyle=true]{csquotes} 

\thesistitle{Medical Concept Normalization in a Low-Resource Setting}
\degree{Master of Science} 
\author{Tim \textsc{Patzelt}} 
\university{University of Potsdam} 
\department{Faculty of Human Sciences} 
\group{Cognitive Systems: Language, Learning and Reasoning} 

\begin{document}


\frontmatter 

\pagestyle{plain} 

\singlespacing

\begin{titlepage}
\begin{center}

\vspace*{.06\textheight}
{\scshape\LARGE \univname\par}\vspace{1.5cm} 
\textsc{\Large Master's Thesis}\\[0.5cm] 

\HRule \\[0.4cm] 
{\huge \bfseries \ttitle\par}\vspace{0.4cm} 
\HRule \\[1.5cm] 
 
\begin{minipage}[t]{0.4\textwidth}
\begin{flushleft} \large
\emph{Author:}\\
\authorname 
\end{flushleft}
\end{minipage}
\begin{minipage}[t]{0.4\textwidth}
\begin{flushright} \large
\emph{Supervisors:} \\
Lisa \textsc{Raithel} \\
Prof. Dr. Manfred \textsc{Stede}

\end{flushright}
\end{minipage}\\[3cm]
 
\vfill

\large \textit{A thesis submitted in fulfillment of the requirements\\ for the degree of \degreename\ in}\\[0.3cm] 
\groupname \\[0.3cm]
\textit{at the}\\[0.3cm]
\deptname\\[2cm] 
 
\vfill

{\large September 6, 2023}\\[4cm] 

\vfill
\end{center}
\end{titlepage}

\begin{abstract}
\addchaptertocentry{Abstract} 
In the field of biomedical natural language processing, medical concept normalization is a crucial task for accurately mapping mentions of concepts to a large knowledge base. However, this task becomes even more challenging in low-resource settings, where limited data and resources are available. In this thesis, I explore the challenges of medical concept normalization in a low-resource setting. Specifically, I investigate the shortcomings of current medical concept normalization methods applied to German lay texts. Since there is no suitable dataset available, a dataset consisting of posts from a German medical online forum is annotated with concepts from the Unified Medical Language System. The experiments demonstrate that multilingual Transformer-based models are able to outperform string similarity methods. The use of contextual information to improve the normalization of lay mentions is also examined, but led to inferior results. Based on the results of the best performing model, I present a systematic error analysis and lay out potential improvements to mitigate frequent errors.
\end{abstract}

\tableofcontents 

\listoffigures 

\listoftables 


\begin{abbreviations}{ll} 

\textbf{ADR} & \textbf{A}dverse \textbf{D}rug \textbf{R}reactions\\
\textbf{BERT} & \textbf{B}idrectional \textbf{E}ncoder \textbf{R}epresentations from \textbf{T}ransformers\\
\textbf{BPE} & \textbf{B}yte-\textbf{P}air \textbf{E}ncoding\\
\textbf{CUI} & \textbf{C}oncept \textbf{U}unique \textbf{I}dentifier\\
\textbf{MCN} & \textbf{M}edical \textbf{C}oncept \textbf{N}ormalization\\
\textbf{NLP} & \textbf{N}atural \textbf{L}anguage \textbf{}Processing\\
\textbf{RoBERTa} & \textbf{R}obustly \textbf{o}ptimized \textbf{BERT} Pre-training \textbf{a}pproach\\
\textbf{SapBERT} & \textbf{S}elf-\textbf{a}ligning \textbf{p}re-trained \textbf{BERT}\\
\textbf{UGT} & \textbf{U}ser-\textbf{G}enerated \textbf{T}ext\\
\textbf{UMLS} & \textbf{U}nified \textbf{M}edical \textbf{L}anguage \textbf{S}ystem\\

\end{abbreviations}


\mainmatter 

\pagestyle{thesis} 

\onehalfspacing
\chapter{Introduction}
\label{section:introduction}

The biomedical field has experienced a surge of data available in digital format in recent years \citep{ohno-machado_natural_2013}. Information about diseases, treatments, medications, and patient experiences are not only proliferating in the scientific literature, but also in health forums, medical blogs, and social media platforms. This richness of data presents opportunities for researchers, healthcare providers, and patients. The opportunities include population analysis, improving health-related search functionalities, and enhancing the understanding of experiences and concerns of patients. However, fully utilizing this vast amount of data necessitates efficient and accurate Natural Language Processing (NLP) methods.

\section{Medical Concept Normalization}

Biomedical NLP is an interdisciplinary field that applies computational methods to biomedical texts. Common data sources are biomedical research published in digital form, electronic health records substitute analog ones \citep{ohno-machado_natural_2013} and the increased usage of social media also produces health-related content \citep{auxier_social_2021}. Biomedical NLP plays a critical role in extracting information from these rich data sources.

Concept normalization, a subset of NLP, is especially relevant in the biomedical domain. It is concerned with the identification and extraction of structured information from unstructured text. The definition of concept normalization is blurry and can vary across researchers and fields. I will use concept normalization as it is defined by \citet{bunescu_using_2006} as the task to link concept mentions to concepts present in a knowledge base. 

The concept mentions appear in their surface form and the target concepts constitute a knowledge base or ontology, often defining relations between the concepts. It is different to named-entity recognition in that the location of concept mentions in a text are already known. 

In this context, this thesis aims to extend the reach of biomedical NLP by applying medical concept normalization to German user-generated texts (UGTs). MCN is a fundamental task in biomedical NLP, aiming to map biomedical expressions in texts to standard concepts in a knowledge base, thus promoting standardization and accessibility. Linking medical concepts from texts written by lay persons to structured medical concepts can help with the following: 
\begin{itemize}
    \item The accuracy and relevance of search results when people are seeking information about health-related topics online can be improved. By identifying the medical concepts mentioned in a particular text, search algorithms can better understand the content and provide more relevant information to users. 
    
    \item Researchers and healthcare professionals can gain new insights into the most commonly discussed health issues and concerns among the general population. This can be especially useful for identifying and responding to outbreaks of infectious diseases. Furthermore, it can be useful for identifying patterns, such as risk factors or adverse drug reactions, in the use of certain medications or treatments. 
\end{itemize}

The primary research question I want to answer in this thesis is: ``How well do deep learning methods normalize German medical lay terms in user-generated texts linked to a large knowledge base?'' The assumption driving this study is that lay terms are harder to normalize and that contextual information can be used to further improve normalization results.
Medical concept normalization (MCN) faces several challenges:
\begin{itemize}
    \item Misleading mentions: Often, many candidate entities share similar surface-level terms with the given mention, making it hard to establish the correct link based only on these terms. Thus, the model must understand the semantics of the mention and its context \citep{vashishth_improving_2021}.

   \item Compound mentions: In the medical field, mentions can be formed of multiple terms. They can be a single word, e.g. \textit{fever}, a compound, e.g. \textit{SARS-COV-2}, or a phrase, e.g. \textit{abnormal retinal vascular development}.

    \item Domain-specific lexicon: Biomedical texts often contain unique terms and abbreviations that are not necessarily present in the training data of common pre-trained language models.
\end{itemize}

Despite the importance of MCN, the current landscape of MCN datasets lacks some essential features. Frequently the type of annotated mentions is under-specified, e.g. with no clear distinctions made between lay and technical terms. Scientific papers or hospital discharge summaries are not further specified with information about the medical sub-field they stem from. Furthermore, only few researchers question how the data they use is obtained and what the annotation guidelines were. This thesis will examine these challenges and propose potential solutions.

One shortcoming of recent work in medical concept normalization, and also in the general NLP research landscape, is the fact that the majority of methods and datasets is based and applied the English language. The focus on English does not only draw a limited picture of the performance of NLP systems, but it also establishes a normative bias and manifests social disadvantages \citep{bender_benderrule_2019}. By working with German medical text, I want to contribute to more diverse, and hence complete, research in NLP. 

Finally, this thesis aims to contribute to the methodological aspects of MCN by addressing the limitations in current evaluation protocols. Existing protocols often overlook a fine-grained error analysis. Evaluation is often conducted in a hit-or-miss fashion, neglecting that medical terms can have shared characteristics and relations that should be considered when evaluating a MCN systems. One example would be the potential ambiguity of medical terms or the proximity of mentions in the knowledge base. In this thesis I want to give a starting point towards a more detailed analysis of the performance of MCN models. I will categorize errors and give suggestions how mitigating some of these errors can be approached.
The code accompanying this thesis can be found under \href{https://github.com/tpatzelt/medical-lay}{https://github.com/tpatzelt/medical-lay}.
\newline

In summary, this thesis extends the scope of medical concept normalization by focusing on the normalization of German medical lay terms in user-generated texts. I hope to promote diversity in biomedical NLP and contribute to the advancements in the field by laying out a systematic error analysis.

\section{Motivating Example}
The goal of this thesis is to implement a medical concept normalization system that receives a text with span markers that indicate where a mention is present in the text. It should assign one concept from a given knowledge base to the mention. The result of the system should look like the following example:  
\begin{center}
    
\begin{table}[h]
\begin{tabular}{|l|l|l|}
\hline
\multicolumn{2}{|l|}{Input}                                                & Output                  \\
\hline
Original Sentence                                      & Mention 
Span(s) & Concept(s)              \\
\hline
\makecell{Seit letzter Woche \textbf{schwitze ich}  \\ \textbf{mich in der Nacht zu Tode}.} & {[}4:11{]}       & \makecell{Sleep hyperhidrosis \\ (C5554276)} \\
\hline
...    & ...       & ...     \\
\hline
\end{tabular}
\end{table}
\end{center}
The mention ``schwitze ich mich in der Nacht zu Tode'' translates to ``sweating to death at night'' in English. It is normalized to the concept \textit{Sleep hyperhidrosis} to which the identifier C5554276 is assigned. The length of the mention and its idiomatic type make it an example that is difficult to assign the correct concept. Furthermore, the example shows the issue of ambiguous mappings. The lay mention could relate to multiple medical concepts in the knowledge base, such as the symptoms of a cold, of hot weather conditions, or of another disease. The lack of a clear 1:1 relationship between lay mentions and medical concepts necessitates a deeper understanding of context and domain-specific knowledge for accurate normalization. The example also demonstrates the difficulty in combining relevant concepts within a mention. Should the focus be on "schwitzen", "Nacht," or should both aspects be considered jointly as a single concept? This adds another layer of complexity to MCN tasks.

\section{Document Structure}
The thesis consists of eight chapters in total. Chapter \ref{section:introduction} provides an overview of the thesis, laying out the context, importance, and objectives of medical concept normalization.
Relevant literature and previous studies in the field of medical concept normalization are presented in Chapter \ref{section:related_work}. The section helps to place the current research in the broader context of MCN and highlights the novel contributions of this thesis.
Next, Chapter \ref{section:background} gives the theoretical and methodological foundations of my thesis. It explains the principles of medical concept normalization and its formal problem definition. I give an overview of the Transformer-based language models that I will use and how these models can be adapted to the medical domain. Moreover, evaluation metrics for MCN and medical knowledge bases will be introduced.
Chapter \ref{section:data_collection} describes the process of annotating an existing dataset with medical identifiers from a knowledge base. It covers how the annotation setup was designed, evaluates the annotation process and depicts the newly generated dataset.
Chapter \ref{section:experiments} details the design and execution of the experiments I conducted. It presents the model choice and the implementation details of the process of model training and validation.
In Chapter \ref{section:results} the outcomes of the experiments are presented. It also provides an evaluation of the models' performance, including an error analysis. Chapter\ \ref{section:discussion} puts the findings into context and possible improvements are sketched out in Section \ref{section:error_mitigation}.
Finally, Chapter \ref{section:conclusion} summarizes the key findings of the research and reflects on the limitations and future outlook of the current study and offers concluding insights.

\chapter{Related Work}
\label{section:related_work}

In this section, I will present the existing literature and studies related to the topic of medical concept normalization (MCN), particularly focusing on the challenges and strategies employed in a low resource setting. First, I examine the datasets commonly used in these studies and their characteristics. Second, the landscape of MCN on English data is explored, presenting an overview of the general methodologies and techniques that have been developed and utilized to address this task. Finally, I will then narrow the focus to studies specifically addressing MCN in low-resource languages. This review of related work provides the necessary context for the research presented in the subsequent sections of this thesis.

\section{Datasets for Medical Concept Normalization}

\begin{table}
\centering
\setlength\tabcolsep{4pt} 
\begin{tabular}{@{} |p{2.2cm}| p{1.7cm}| p{3cm}| p{2.4cm}| p{2.5cm}|@{}}
\toprule
\textbf{Dataset} & \textbf{Language} & \textbf{Source} & \textbf{Size} &  \textbf{Annotations} \\
\midrule
COMETA & English & Medical Reddit & 20.000 mentions & SNOMED-CT  \\
\hline
CADEC & English & Medical online forum (ADR) & 1.253 posts & SNOMED-CT, MedDRA \\
\hline
PsyTAR & English & Medical online forum (ADR) & 6.009 posts & SNOMED-CT \\
\hline
SMM4H 2022 & English & Medical Twitter (ADR) & 6.556 mention & MedDRA \\
\hline
MedMentions & English & Biomedical research papers & 4.392 abstracts & UMLS \\
\hline
AskAPatient & English & Medical online forum &  8.662 mentions & SNOMED-CT, MedDRA \\
\hline
TwADR-S & English & Medical Twitter (ADR) & 201 mentions & SNOMED-CT \\
\hline
TwADR-L & English & Medical Twitter (ADR) & 1.436 mentions & SNOMED-CT \\
\midrule
QUAERO & French & Biomedical research papers & 26.409 mentions & UMLS Semantic Groups \\
\hline
MANTRA & Multi (EN, FR, DE, ES, NL) & Biomedical research papers & 5.530 mentions & Subset of UMLS \\
\hline
TLC & German & Online patient forum (kidney, stomach) & 7.390 mentions & Medical and lay terms with synonyms\\
\midrule
NCBI Disease Corpus & English & 793 Biomedical research papers & 793 abstracts &  MeSH \\
\hline
BC5CDR Corpus & English & Biomedical research papers & 1.500 abstracts & MeSH\\
\hline
i2b2 Obesity Challenge Dataset & English & Discharge summaries & 1.237 records & Obesity Comorbidities \\
\bottomrule
\end{tabular}
\caption{Datasets for medical concept normalization.}
\label{tab:MCNDatasets}
\end{table}

In this section, I will turn towards the various datasets utilized for MCN. Datasets form the backbone of any machine learning task, and in the context of MCN, they can be diverse, ranging from scientific biomedical texts, electronic health records, to lay person texts derived from social media or online forums. An emphasis is put on datasets that are available in languages other than English. An overview of the different datasets and their characteristics can be found in Table \ref{tab:MCNDatasets}. 
I will put particular focus on datasets that are annotated with UMLS concepts. The Unified Medical Language System (UMLS) is a comprehensive set of multiple healthcare vocabularies and standards, aiming to improve data interoperability across different healthcare systems. It is the largest unified source of medical concepts with over 4 million entries. A detailed description is provided in Section \ref{section:ums}.

To start off with English datasets, COMETA \citep{basaldella_cometa_2020} stands out with 20.000 mentions from English Reddit medical forums, annotated with \\SNOMED-CT, a controlled vocabulary in UMLS. Similarly, CADEC \citep{karimi_cadec_2015} is another English dataset composed of medical forum posts. The posts are annotated with Adverse Drug Reactions (ADRs), Drug, Disease and Symptom labels from SNOMED-CT. The PsyTAR \citep{zolnoori_psytar_2019} and Social Media Mining for Health \citep{weissenbacher_overview_2022}  dataset also provide ADR annotations from patient narratives in medical forums and Twitter, respectively. Moreover, MedMentions \citep{mohan_medmentions_2019} comprises a large corpus of 4.392 biomedical research papers annotated with UMLS, while AskAPatient \citep{limsopatham_normalising_2016}, TwADR-S and TwADR-L \citep{limsopatham_adapting_2015} all focus on mapping medical forum and Twitter posts to medical concepts from UMLS.

Apart from English datasets, QUAERO \citep{neveol_quaero_2014} and MANTRA \citep{kors_multilingual_2015} provide non-English alternatives. QUAERO focuses on French biomedical texts, annotating ten entity categories corresponding to UMLS Semantic Groups.

MANTRA, on the other hand, provides text units from different parallel corpora in English, French, German, Spanish, and Dutch, annotated with a subset of UMLS.

The focus of my work, however, is on German medical concept normalization, which makes the Technical-Laymen Corpus (TLC) \citep{seiffe_witchs_2020} particularly relevant. This dataset contains German forum data from Med1.de, an online patient forum, where users talk about health, life help and well-being. The dataset is built from the sub-forums focused on kidney diseases and stomach and intestines only. Each forum post contains annotations that indicate whether a mention is a technical or lay medical term. The span of each mention is annotated along with an synonym for it. The annotators were requested to provide a technical synonym if the mention is a lay term and vice versa. It provides a valuable foundation for my research, given the language and the focus on patient language in online forums. However, it lacks UMLS annotation which will be added during the course of this work. The initial annotations are useful for the task of entity detection, where a system should determine the location of medical mentions in a text, or for named entity detection, where a system should find medical mentions and assign them either as lay or technical. In this work I want to explore medical concept normalization which assumes the spans of medical mentions to be known and assigns, based on the spans, a normalized concept from knowledge base to each mention.  For this task further annotations are necessary.

Several other datasets, though not explicitly designed for medical concept normalization, provide valuable resources that can be repurposed for this task. 

The NCBI Disease Corpus \citep{dogan_ncbi_2014}, consisting of PubMed abstracts annotated with disease mentions and related concepts, offers a rich ground for learning and testing medical concept normalization methods.

Similarly, the BC5CDR Corpus \citep{li_biocreative_2016}, annotated for disease named entity recognition and chemical-induced disease relation extraction, can be exploited for mapping informal or alternative disease and chemical names to their standard forms. Finally, the i2b2 Obesity Challenge Dataset \citep{uzuner_recognizing_2009}, comprising a set of 1.237 discharge summaries focused on obesity and diabetes, offers real-world patient narratives that could be used to train and evaluate models on the task of normalizing diverse medical expressions. 

In summary, numerous datasets are available for medical concept normalization, each covering a different variety of types of data. The majority of these datasets are developed based on English texts. They include data from multiple sources such as medical forum posts, social media, discharge summaries, and medical research papers. These types of data differ in style. For instance, forum posts and social media content usually contain informal and colloquial language, while discharge summaries and research papers incorporate formal and technical medical terminologies.
However, there are also non-English datasets, like QUAERO, MANTRA, and the notable TLC for German. These datasets serve as central resources to evaluate the performance of medical concept normalization models on non-English data. Similar to their English counterparts, these datasets also include diverse data types such as medical forum discussions, social media content, and formal medical reports. While the TLC dataset is not the only one covering mainly lay language, it is the only dataset that covers German lay language.

\section{Medical Concept Normalization Methods}
\label{section:related_mcn_methods}
Over the last decade, the task of medical concept normalization has gained significant attention and interest. MCN aims to overcome the challenges associated with multiple terminologies and vocabularies used to represent the same medical concept, ensuring that healthcare providers, researchers, and other stakeholders have systematic access to information from various data sources. 
\newline
Among the first systems, cTAKES \citep{savova_mayo_2010} and MetaMap \citep{aronson_overview_2010} were rule-based systems that use lexical, morphological and syntactic features to detect and normalize concept mentions. 

cTAKES aims to provide a component-based end-to-end system that uses a feature set of both lexical and contextual features, e.g., part-of-speech tags and surrounding words, to normalize medical entities. The normalization component is dictionary-based, but tries to boost recall by applying lexical permutation to the entries in the target dictionary.

MetaMap is set up in a similar way. It uses a combination of lexical, syntactic, and semantic information to normalize medical entities. A dictionary lookup is used with a scoring algorithm that takes into account various features such as the frequency of occurrence of each candidate concept in the UMLS Metathesaurus, its semantic type(s), and its distance from other concepts in the input text. The highest-scoring concept is then selected as the most likely target concept for a mention. 

The authors of cTAKES and MetaMap both do not provide an quantitative analysis of their systems. The work of \citet{reategui_comparison_2018} revisits the systems and lays out a fine-grained analysis of their performance on a subset of the i2b2 Obesity Challenge data \citep{uzuner_recognizing_2009} containing 1.237 discharge summaries. All manually annotated mentions that belong to one of 14 obesity comorbidity entities were used for evaluation. Negated mentions or mentions containing typographical errors are excluded. As result, cTAKES slightly outperforms MetaMap with an F\textsubscript{1}-score of 0.89 over 0.88. The authors conclude that a combination of both systems yields the best results and hard-coded components like the internal list of abbreviations for expansion can still be extended. 
\newline
A conceptually simple approach to MCN is approximate string matching. In the most straight forward way, the concept names in a knowledge base can be searched for a string representation of a mention. This approach suffers the problem that concepts are often referred to by synonyms, hypernyms or hyponyms which can only be disambiguated by the context they appear in. Furthermore, user-generated texts can differ largely in structure, style and  do not have to be grammatically correct and can contain typos. Consequently, the recall of such systems can be fairly low. 

In this paragraph, I will introduce string-based similarity methods and their applications in medical concept normalization. QuickUMLS, developed by \citet{soldaini_quickumls_2016}, is an efficient concept normalization system that parses mentions in unstructured text into tri-grams and leverages a similarity measure to identify matching concepts. The approach tokenizes the document and generates valid sequences of tokens within a specified window size, ensuring they do not span across sentences, start with punctuation, or contain only stop words or numbers. It identifies strings in the dictionary that share a specified number of features with the target string. The mention and target strings are represented as sets of features, e.g., character trigrams, and constitute an index used to associate each feature with the strings containing it. The system then identifies strings in the dictionary that have more than a specified number of features in common with the target string. QuickUMLS has been shown to slightly outperform cTAKES and MetaMap on a dataset of 1.254 clinical reports, while being on average 25 times faster.

The main contribution by \citet{seiffe_witchs_2020} is the TLC dataset that comprises user-generated posts from an German online medical patient forum. But the authors also present a baseline for MCN: They enrich each concept in a subset of UMLS with more synonyms from a thesaurus, in that case Wiktionary. Then they use simple string search to find the synonyms of each concept in the TLC dataset. By doing so, they are able to normalize $\sim$64\% of all lay and technical mentions in the dataset. In comparison, simple string search without the additional synonyms from Wiktionary resulted in only $\sim$57\% of mentions being found. The exact method is further described in Section \ref{section:baseline}.  

The first data-driven approach for MCN was implemented by DNorm \citep{leaman_dnorm_2013} using a pairwise learning-to-rank approach. Mentions and concept names are represented as Term Frequency-Inverse Document Frequency vectors. The Term Frequency measures the occurrences of a token in a mention or name, while the Inverse Document Frequency is calculated based on the number of names containing the token. A learned weight matrix contains the correlation between tokens, allowing the model to represent positive and negative correlations, as well as synonymy and polysemy. The goal is to learn a scoring function that gives higher scores to matching pairs than to mismatched pairs. The model iterates through all possible concept names and ranks them based on the calculated score. For evaluation the authors use the NCBI disease corpus \citep{dogan_ncbi_2014}, which contains 793 PubMed abstracts. MetaMap and DNorm reach a F\textsubscript{1}-score of 0.56 and 0.80, respectively. Furthermore, the authors of DNorm also proposed TaggerOne \citep{leaman_taggerone_2016}. For normalization they use a semi-Markov structured linear classifier. It is applied to embeddings of mentions and concepts names that are learned using orthographic, morphological, and contextual features. TaggerOne achieves am improvement over DNorm of $\sim$0.2 absolute points of F\textsubscript{1}-score on the NCBI disease and the BioCreative 5 CDR \citep{li_biocreative_2016} dataset. 
\\
With the rapid advancements in deep learning, there has been a significant shift towards employing methods utilizing artificial neural networks for MCN. Early years of Deep Learning saw the development of simple neural models like feed-forward and recurrent neural networks, which were successful in tasks like sentiment analysis and machine translation. The evolution of neural networks for natural language processing proceeded with the development of complex models like Convolutional Neural Networks and Long Short-Term Memory Networks \citep[p. 439-442]{lecun_deep_2015}, which were successful in capturing long-term dependencies and patterns in language. The success of these methods can be attributed to their ability to learn effective representations of text data and to model the sequential nature of language. The rise of Transformers, with the introduction of the attention mechanism \citep{vaswani_attention_2017}, paved the way for the emergence of models like Bidirectional Encoder Representations for Transformers (BERT) \citep{devlin_bert_2019} and Generative Pre-trained Transformers (GPT) \citep{radford_improving_2018}, which are successful in capturing contextual information. These methods can derive meaningful features from the data they are trained on and show strong performance in capturing complex patterns, hierarchical relationships, and context-specific features. These deep learning methods hold significant potential for application in specialized fields such as medical concept normalization, enabling the understanding of complex medical terminologies from diverse data sources.
\\
Deep learning-based methods for medical concept normalization can be categorized into classification, generate and rank, and embedding similarity models. Classification models categorize medical terms into predefined classes, generate-and-rank models produce and order potential normalized concept candidates based on their likelihood, and embedding similarity models finding the closest vector in the space of concept embeddings for normalization. While deep learning encompasses a wide array of objectives, these three stand out as particularly prominent in the field of MCN. The following three sections will describe each of them.

\subsection{Classification}
Medical concept normalization can be framed as a supervised classification task for deep learning models by representing text as embeddings, convert target concepts from the knowledge base into numerical labels, and training a deep learning model to map the input embeddings to the corresponding numerical labels. This approach may struggle with a high number of output classes due to a weak training signal and potential over-fitting. The performance of the model can be evaluated using standard metrics such as accuracy, precision, recall, and F\textsubscript{1}-score. \cite{limsopatham_normalising_2016} compare two approaches to assign medical concept to social media messages. They use static pre-trained word embeddings that are processed by either a Convolutional or a Recurrent Neural Network. The convolution operation is performed on representation of each sentence in the input. In contrast, the Recurrent Neural Network processes the word representations sequentially from left to right. Both methods produce a fixed sized embedding that is passed through a max pooling layer and then used as input to a softmax layer for multi-class classification. 

Building upon the work of \citet{limsopatham_normalising_2016}, \citet{miftahutdinov_deep_2019} investigate how prior knowledge from UMLS can be combined with a learned representation that captures contextual semantic information of an entity mention. This is achieved by fusing the representation of an entity mention and concepts from UMLS. Concatenating the entity mention representations with similarity features yielded a performance gain of 2-5\% on the respective test set of the CADEC \citep{karimi_cadec_2015}, PsyTar \citep{zolnoori_psytar_2019} and SMM4H \citep{weissenbacher_overview_2022} dataset.

\citet{li_fine-tuning_2019} proposed to first fine-tune different Transformer-based models on a large collection of electronic health records with the next word prediction training objective. Then they use the classification token representation to predict the class of mentions in three different datasets, as it was done by \citet{limsopatham_normalising_2016, miftahutdinov_deep_2019}. The number of classes in each dataset ranged from 11.000 to 380.000. They show that self-supervised fine-tuning on medical data slightly improves classification performance. Furthermore they state that classification performance decreases with number of classes, i.e. all models reach an F\textsubscript{1}-score of $\sim$0.9 on the datasets with $\sim$11.000 classes and only $\sim$0.4 F\textsubscript{1}-score on the dataset with 380.000 classes. 

One approach to tackle the problem with a high number of classes is to employ a type prediction component which implicitly narrows down the number of plausible concepts for a mention. In UMLS each concept is also annotated with one of 127 semantic types, which constrains the number of possible target concepts for a mention by over two order of magnitudes. One example of this approach is presented by \citet{zhu_latte_2020}. They used static word and character embeddings of a Convolutional Neural Network fed to a Long Short-Term Memory Network to predict the semantic type of a mention. They jointly trained a classifier to predict types and to rank the concept predictions for each mention. The combined model can utilize semantic type information for the mention classification. The version of their model without known type modeling under-performs compared to their full model. This implies that multitasking with type classification greatly improves concept normalization.

Similarly, \citet{vashishth_improving_2021} introduced a Transformer-based semantic type prediction module, that can be used to reduce the number of concept candidates for mention classification. Incorporating their new model into existing systems, i.e. cTAKES, MetaMap and QuickUMLS, provides a measurable improvement, especially in the recall of the systems.

For the BioCreative VII Task 3, \citet{weissenbacher_biocreative_2021}, \citet{roller_boosting_2021} applied a Transformer-based model, combined with string matching using background knowledge, to an highly imbalanced dataset of tweets containing drug mentions. The background knowledge was extracted from the SMM4H dataset \citep{sarker_overview_2017} in the form of mention and target concept pairs. To further increase the performance of the Transformer-based model, the authors applied different text data augmentation techniques \citep{wei_eda_2019}, added additional training data from SMM4H dataset and generate new training data by substituting new entities into existing samples.

\subsection{Generate and Rank}
The Generate and Rank approach for medical concept normalization first generates normalized candidate concepts for a given input mention, using simple dictionary look-ups or advanced machine learning methods that output a confidence score. After potential matches are found, a ranking model scores and ranks these candidates based on factors such as semantic similarity to the original mention and its contextual usage. The top-ranked candidate is then chosen as the final normalized concept. This approach is flexible and adaptable because it can be decomposed into two models, but consequently it requires substantially more computational resources than a single model. 

\citet{li_cnn-based_2017} introduce a convolution-based ranking method that first generates candidates using ten handcrafted rules, and then ranks the candidates according to their semantic information modeled by Convolutional Neural Network as well as their morphological information. The semantic information is obtained from word2vec embeddings \citep{mikolov_distributed_2013} and the morphological information is based on string similarity. The advantage of this approach is that it is based on a trainable neural network and learns the distribution of the concepts in the data, which leads to better performance than traditional rule-based methods. Experiments on two datasets \citep{dogan_ncbi_2014, suominen_overview_2013} show that the proposed ranking method outperforms traditional rule-based method. 

Extending simple similarity-based predictions, \citet{ji_bert-based_2020} generate concept candidates by using a probabilistic string similarity framework \citep{robertson_probabilistic_2009}. They fine-tune different language models to rank the candidates with a sentence-pair classification task. The sentence-pair classification works by inserting a separation token between the mention and each target concept candidates. The separation token of each sentence pair was used as input to a binary classifier that predicted the probability of each target concept belonging to the input mention. 

In very similar way, \citet{xu_generate-and-rank_2020} design their system, but replace string similarity-based candidate generation by a Transformer-based model. Their novel two-step approach leverages BERT for both generation and ranking. They also introduce a semantic type regularizer to improve rankings by considering semantic types. Their evaluation include many MCN models and their final model achieved the best results on two social media test sets, TwADR-L and AskAPatient. The string-based approach for candidate generation and the Transformer-based approach for candidate ranking reaches the best results on the clinical MCN test set \citep{luo_mcn_2019}. The addition of the Transformer-based ranker and the semantic type regularizer consistently improves performance across test sets, with the most substantial gains observed in the MCN dataset, indicating that the regularizer is particularly effective at managing large candidate concepts and unseen concepts during training. 

Furthermore, \citet{mondal_medical_2019} used a Convolutional Neural Network with triplet loss to rank the candidates obtained by similarity measures on static but domain-specific word embeddings. Using Triplet Loss can improve learning mention and concept representations in settings where positive samples are expensive and negative samples are cheap to obtain, as it is often the case in MCN. The Triplet Loss is a loss function that minimizes the distance between a concept mention and its positive target concept, while maximizing the distance between the mention and its negative target concept. It aims to learn a feature space where mentions of the same class are closer together, while mentions of different classes are further apart. The models shows a very strong accuracy of 0.9 on the NCBI dataset. 

\citet{hristov_application_2021} fine-tune a BERT model for multi-cluster classification where clusters are obtained beforehand by k-means clustering on concept embeddings. After a mention is assigned to one or more clusters, a one-vs-rest Support Vector Classifier predicts the concrete concept of the mentions.

In addition, an efficient method for MCN is proposed by \citet{abdurxit_efficient_2022}. They model relations between mentions and concepts themselves and between each other by an inter- and intra-attention neural network. The representations of the mention and each concept are then aggregated in a convolutional layer and scores for each concept are generated by applying a softmax layer. Using a special attention mechanism maintains a similar performance to comparable models while maintaining fewer parameters and higher inference speed.

\subsection{Embedding Similarity}
A different approach to MCN is to produce mention embeddings and then find the closest concept embedding in the embedding space. In this approach, both the mention and the target concepts are transformed into high-dimensional vector embeddings that capture the underlying semantic meaning of the input. These embeddings can then be compared using techniques such as cosine similarity (Section \ref{section:cosine_similarity}) to determine the degree of similarity between the inputs. It differs to classification in that the target concept is not encoded by a class label, but by the vector embedding of that concept. The method for generating the vector embedding of the target concept does not necessarily have to be same as for generating the mention representation, but usually this is the case. A fine-grained analysis of the target embedding space can also reveal insights into how the target concepts relate to each other, e.g., by clustering them or using a visualization technique based on dimension reduction. Such an analysis is not possible with classification or the generate-and-rank method. Furthermore, the embedding step can be separated from the candidate matching step. The advantage would be that the concept embeddings can be cached, leading to faster inference time.
\newline

\citet{phan_robust_2019} use skip-gram mention and concept embeddings and refine them with a Long Short-Term Memory Network trained on a synonym-, \\ concept-, and context-based training objective. The target embedding with the closest cosine similarity is chosen as target concept. They show that using robust mentions embeddings can be used effectively for medical concept normalization. Robust embeddings are defined as embeddings of mentions that are similar to its synonyms as well as its conceptual and contextual representations.

A comparison between an approximate string matching method to neural embeddings with linear transformation alignment is presented by \citet{basaldella_cometa_2020}. As neural embeddings they use static and Transformer-based embeddings. A linear alignment function is trained to find the closest concept by using Triplet Loss on triplets of the mention, positive and negative target concept. They conclude that string-based methods are a solid baseline, but neural embeddings are better suited for MCN because they have access to context information. In line with this insight, most recent systems realizing the similarity approach use Transformer-based embeddings and fine-tune them so that mention and concept embeddings are close to each other in the embedding space. 

In an exploration of Transformer-based embeddings, \citet{kalyan_medical_2020} devised an approach that leverages a Robustly Optimized BERT Pretraining Approach (RoBERTa) model \citep{lee_biobert_2020} to generate mention and concept embeddings. By employing cosine similarity, the model identifies the closest target concept for a given mention. This straightforward method achieved impressive results on three standard datasets (CADEC, PsyTAR, and SMM4H) surpassing existing methods with improvements of $\sim$2.31, $\sim$1.26, and $\sim$1.2 in accuracy, respectively. Remarkably, these performance gains were realized without the need for augmenting the training set with synonyms, a testament to the efficacy of contextualized word embeddings. 

Building on this simple embedding similarity approach, \citet{miftahutdinov_medical_2021} introduced an enhancement involving the Triplet Loss with specific sample selection methods. Their first method provides a diverse training sample set by selecting several concept names with the same identifier as positive examples and random concepts as negatives. To further enrich the model's training, the second method incorporates concept names from a concept's hierarchical parents. The third method, after training a model with random sampling, identifies both positive and difficult negative examples, which are most similar but incorrect. Lastly, the fourth method adapts the third by introducing concept names from a concept's siblings as negatives, broadening the model's understanding of concept relationships.

Further innovation came from \citet{sung_biomedical_2020}, who introduced BIOSYN, a model that employs both sparse, based on Term Frequency, and dense, based on BERT embeddings, representations for mentions and synonyms, utilizing a shared encoder. To output a similarity score, the similarity function combines these representations using a trainable weight. The model is trained to maximize the marginal probability of positive synonyms among the top candidate concepts for a mention, which are updated iteratively. During inference, the model identifies the nearest synonym to an input mention using the dot product instead of the widely used cosine similarity.
\newline

The Transformer-based embedding methods presented before do not use external knowledge about the ontology constituted of the target concepts. Several methods exist to leverage knowledge from the UMLS knowledge graph into pre-trained language models. 

\citet{yuan_improving_2021} use a Transformer-based encoder to produce embeddings of the concepts in UMLS. They calculate a distance matrix of concepts where the distance between two concepts is based on how many edges they are apart in the UMLS knowledge graph. The distance matrix is fused into the embedding layer of a separate Transformer-based encoder that produces the mention embeddings. As fusing method they use summation. 

The work of \citet{zhang_knowledge-rich_2022} shows how to generate data suitable for self-supervision from unlabeled biomedical text. For each concept mention in UMLS, they collect text samples from a large dataset of biomedical publications that contain the concept mention. To find the right samples, they use string matching which does not have high recall but, more importantly, does have a high precision. Their embedding model is trained with Contrastive Loss so that the contextual embeddings for samples that contain a mention of the same concept are pushed together and unrelated samples are pushed apart. In the scenario where labeled examples are available, the method can directly use them in a ``lazy learning'' approach without additional training. This allows gold mention examples from target training data to be used as mention prototypes for normalization, augmenting the self-supervised ones. 

\citet{liang_fast_2022} propose a very efficient MCN model that is pre-trained using the Continuous Bag-of-Words model, which significantly improves both efficiency and performance. Consequently, the model achieves comparable performance to most baseline methods in only 30 seconds of training and is 3000-5600 times faster during inference. For finding the most similar concept for a given input mention, they also use the maximum cosine similarity.

Another method to find the most similar concept embedding for a mention is to use a cross-encoder network. A cross-encoder is a neural network that produces for a source text a target-dependent source embedding using a source-target pair. 

\citet{lin_enhancing_2022} use such an cross-encoder to produce mention embeddings that are dependent on the concept it is used as input with. A problem arising from the high number of concepts is the quadratic time complexity of the cross-encoder because it needs to be computed for each mention concept pair. To mitigate this, a reduced candidate list is generated using fast cosine similarity ranking. The cross-encoder is then applied to re-rank these candidates. 

One of the most promising approaches to leverage UMLS knowledge is the approach sketched by \citet{liu_self-alignment_2021}. They describe a metric learning framework that learns to self-align synonymous biomedical entities. The framework can be used as both pre-training on UMLS, and fine-tuning on task-specific datasets. It does not need more than an aligned list of all concept names and identifiers in UMLS. Inference can be implemented by simple nearest neighbour search. The authors present SAPBERT, a self-alignment pre-training scheme for learning biomedical entity representations. It works by combining Triplet Loss with a technique to find the most discriminative triplet of a concept name, a positive and a negative concept name. They highlight the consistent performance boost on many biomedical downstream tasks. BIOSYN combined with the SAPBERT pre-training marks the current state-of-the-art on the biomedical scientific datasets NCBI, BC5CDR and MedMentions. For the biomedical social media language datasets COMETA and AskAPatient, further pre-training of PubMedBERT \citep{gu_domain-specific_2021} with SAPBERT and using nearest-neighbour search is the current state-of-the-art.
The drawback of fine-tuning BERT-based models is the computational demand of such models. Addressing this problem, \citet{lai_bert_2021} propose to use a Residual Convolution Neural Network \citep{he_deep_2016}, aiming to capture local interactions efficiently. They use the first embedding layer of PubMedBERT \citep{gu_domain-specific_2021} to create initial vector representations of input mentions. The model's encoding layer consists of convolutional filters, activation functions, and residual connections to mitigate the vanishing gradient problem. The final vector representation is obtained via max pooling or self-attention pooling. They use ten times less model parameters than a typical language model while still achieving results comparable to the state-of-the-art models (SapBERT, BIOSYN) on four datasets (MedMentions, COMETA, BioCreative V CDR, NCBI). 

\section{Medical Concept Normalization for Languages other than English}

In this section, I will show MCN methods specifically tailored to operate on non-English datasets. While considerable work has been done on English data, extending these methodologies to other languages poses unique challenges. Key among these are the dominance of English names in UMLS, and the scarcity of non-English pre-trained models and labeled data. I will discuss these challenges and introduce methods that have been developed to mitigate these issues and enable effective MCN in non-English languages.  

The work of \citet{liu_learning_2021} investigates the challenge of injecting domain-specific knowledge, such as that available from UMLS, into pre-trained language models to handle languages other than English. They benchmark their system on a set of ten diverse languages. The authors demonstrate that standard non-English monolingual and multilingual language models fall short in performing this task compared to their English counterparts. They propose a series of cross-lingual transfer methods to transfer domain-specific knowledge from resource-rich languages like English to those that lack such resources. The methodology includes creating similar representations for synonyms across different languages during the model training, based on the SAPBERT method proposed by the same authors. The learning scheme is composed of an online sampling procedure for training example selection, and a metric learning loss that encourages synonyms sharing the same concept identifier to obtain similar representations. They introduce a state-of-the-art multilingual encoder that improves performance in cross-lingual MCN, especially for languages with fewer resources.

Another example of injecting cross-lingual knowledge into general pre-training language models is shown by \cite{lin_enhancing_2022}. Instead of a vanilla Transformer-based architecture, they used a Sentence-BERT model \citep{reimers_sentence-bert_2019}. A Sentence-BERT model consists of two weight-sharing BERT Models and is trained on sentence pairs instead of single sentences. They demonstrated that Sentence-BERT models, when injected with a relevant subset of UMLS, can perform comparably or even slightly better than models pre-trained on biomedical corpora. As relevant subset they selected three controlled vocabularies from UMLS that cover more than 90\% of all concepts. Injecting only a relevant subset of UMLS yielded better results using the full UMLS ontology while also being faster to use.

\citet{frei_annotated_2022} propose a method to leverage pre-trained language models for training data acquisition, thereby enabling the creation of large datasets to train smaller, more efficient, and task-specific models. They demonstrate their approach by creating a custom dataset used to train a medical Named Entity Recognition model for German texts, named GPTNERMED. They emphasize that their method is fundamentally language-independent. This was achieved by leveraging the model's few-shot learning capability. They introduced a few annotated text samples in a simple markup format to the model, which then generated new, similar sentences by autoregressively predicting the next word based on the provided examples. This resulted in a synthetic corpus with corresponding annotations. The generated data was then cleaned and used to fine-tune a pre-trained BERT encoder models for the MCN task. This process proved to be effective in creating a domain-specific dataset without requiring extensive manual annotation.

For an analysis of cross-lingual MCN performance, \cite{schwarz_multilingual_2022} use pre-trained language models, PubMedBERT \citep{gu_domain-specific_2021} and XLM-RoBERTa \citep{conneau_unsupervised_2020}, to generate embeddings of medical terms. Synonyms, including those in different languages, are positioned closely in the embedding space. Similarity between embeddings is determined using Euclidean distance. If the Euclidean distance exceeds a threshold, the term is broken down into smaller parts to identify sub-spans for normalization. Their model is able to normalize mentions from a different language to UMLS as long as the source was part of the pre-training. However, inherently it struggles with normalizing entities that share a name but have different meanings based on context.
The authors of \citep{yuan_coder_2022} propose CODER, a novel approach for biomedical concept normalization using knowledge graph-based contrastive learning. They introduce a method for generating medical term embeddings that incorporates both synonyms and, in contrast to SAPBERT, relations from the UMLS knowledge graph. They include synonyms from languages other than English in the fine-tuning process and evaluate the performance on 14 different languages. They find that normalization works well on letter-based languages, like English and German, but fails to assign target concepts for mention from character-based languages like Chinese and Japanese.

\citet{alekseev_medical_2022} do not propose a new method but critically discuss how current models are often evaluated in narrow and biased settings, potentially leading to overestimated performance. In response to this, they introduce a more sophisticated evaluation approach, combining cross-terminology and cross-lingual testing on real-life biomedical and clinical texts. The authors question the performance overestimation of current benchmarks in English, Spanish, French, German, and Dutch. They also investigate the ability of a model trained in English to generalize for zero-shot MCN in other languages. Furthermore, they evaluate different word representation types for cross-lingual clinical entity linking. Their results emphasize the importance of test set filtering to prevent data leaks for a fair evaluation. Filtering test sets to avoid leaks refers to the process of removing any data from the test set that is identical or highly similar to data in the training set. This is done to ensure that the model performance is evaluated on completely unseen data, thereby providing a more accurate measure of its generalizability. In their evaluation protocol, they find that sparse baselines often outperform BERT-based models and that domain knowledge significantly improves the results. They also note that fine-tuning on medical datasets yields improvements that common benchmarks without filtering do not show.

\chapter{Background}
\label{section:background}
In this section I want to introduce and explain all the methods and knowledge bases that I used throughout this thesis. The following content is not intended as exhaustive description of the methods that are available but rather focuses only on the methods I used. These are mainly the Unified Medical Language System, word embeddings, Transformer-based models and Sentence Cross-Encoders and a formal problem definition for MCN.

\section{Problem Definition}

Formally, for a given mention $m$, the context $ctx_a, ctx_b$ before and after the mention and a knowledge base containing $N$ concepts $\mathcal{C}=\{c_1,c_2,...,c_n\}$, the task is to find concept $c_i \in \mathcal{C}$ that $m$ refers to. Each $c \in \mathcal{C}$ is defined by a unique identifier. A knowledge base contains also primary names, also called the \textit{Preferred Terms}, and a list of alternative names. The set of all primary and alternative names in the knowledge base is $\mathcal{N} =\{n_1,n_2,...,n_m\}$ where $m$ is the number of names. A knowledge base can then be seen as surjective function $f: \mathcal{N} \to \mathcal{C}$ that assigns to each concept $c$ at least one name $n$. Putting it all together, a MCN model can be defined as function $$g: (ctx_a, m, ctx_b) \to c $$ that maps a mention $m$ with its optional context $(ctx_a, ctx_b)$ to one concept $c \in \mathcal{C}$ from the knowledge base.

\section{The Unified Medical Language System}
\label{section:ums}

The Unified Medical Language System (UMLS) \citep{bodenreider_unified_2004} is a compendium of many controlled vocabularies in the biomedical sciences and provides a unified access to concepts and relations that are indexed in different terminologies. It is created and maintained by the U.S. National Library of Medicine.

The UMLS consists of three primary components:
\begin{itemize}
    \item Metathesaurus: This is the largest component and contains information about biomedical concepts, their synonyms and their relationships to one another. It integrates over 200 different biomedical vocabularies and classifications.

    \item Semantic Network: It provides a graph-based categorization for all concepts represented in the UMLS Metathesaurus. The network defines a set of broad categories, known as semantic types, and a set of useful relationships, known as semantic relations, between these types. These relationships are of various types, such as synonymy, hierarchical, associative, and temporal.

    \item Specialist Lexicon and Lexical Tools: This component provides lexical information about biomedical and general English words and provides a software tool to process medical texts.
\end{itemize}

The primary goal of UMLS is to improve interoperability among biomedical information systems by addressing the variability in the language of biomedical information. This includes the use of different names in the same or different language for the same concept and a set of relations that can exist between concepts. This strength makes UMLS a good choice for medical concept normalization, because data sources annotated with different vocabularies can be jointly used.

The UMLS Metathesaurus contains information about more than 4 million biomedical concepts, which are drawn from over 200 different sources or vocabularies. One concept consists of a concept name, a concept unique identifier (CUI), a description, a list of identifiers for the concept in other vocabularies and a list of synonyms. These concepts are linked by over 14 million relationships. The number of concept names and synonyms is over 13 million.

A new version of UMLS is published twice a year. A new release can contain retired CUIs which are caused by either a removal of a concept or if two concept are merged into one.

\section{Mapping Words to Numbers}
NLP models, and computers in general, operate on numerical representations, such as numbers, vectors or matrices. However, natural language operates in a different space. It is composed of words, which in turn can be defined as a sequence of letters and symbols. This section gives a definition of word embeddings and how different embedding instances can be compared. 

\subsection{Word Embeddings}
Word embeddings are a fundamental concept in NLP that is important to understand before explaining the models discussed in this thesis. 
For computational processing, words or segments known as tokens (refer to Section \ref{section:tokenization}), must be translated into numerical forms, ideally vectors. Thus, a word embedding can succinctly be described as a word's vector representation. In the following, I will refer to word embeddings as embeddings, as they do not necessarily represent words but rather tokens. The vector representation can be used in downstream tasks, e.g., to assign a class to a sequence of embeddings. The primary objective of an embedding model is to optimize these such that they are useful in downstream tasks. 

An embedding model consists of two parts: The \textit{vocabulary} assigns to each possible token an integer ID. The choice of IDs can be random. The \textit{embedding model} is the model that creates meaningful representations from each token ID. It can be a simple neural network or an advanced architecture like BERT \citep{devlin_bert_2019}. The size of the resulting representation for each token depends on the model and for BERT-based model is typically set to either 512 or 1024. 

There is an important difference between token IDs and token embeddings that I want to stress. The IDs do not carry any meaning because they are assigned randomly. In modern NLP, the meaning is based on the Distributional Hypothesis by \citet[p.11]{firth_synopsis_1957}. It contains the famous quote ``You shall know a word by the company it keeps'', stating that the meaning of a word is defined by the context is appears in. The vocabulary assigns IDs to single tokens irrespective of the context. In contrast, the embedding model considers the whole input sequence, including each value in it but potentially also the position of each value, and can produce a so called \textit{dynamic contextualized embedding}. 

Consider the vocabulary $V = \{I: 1,\ am: 2 ,\ happy: 3,\ sad: 4 \}$ and the two sentences $x_1=I\ am\ happy$ and $x_2=I\ am\ sad$. Then, $V(x_1)=[1,2,3]$ and $V(x_2)=[1,2,4]$. Applying an contextualized embedding model to both vector representations would yield two embeddings that have different values in the first two columns, although the IDs, 1 and 2, are the same. This is because they appear in a different context, i.e. with the values 3 and 4 as the last element.

\subsection{Cosine Similarity}
\label{section:cosine_similarity}

Before covering models that can be used to create embeddings, it makes sense to explain what insights we can get by comparing embeddings of different words. 
The ability to compare embeddings brings significant insight into their semantic relatedness. Specifically, it allows to quantify the degree to which different words or phrases share contextual information according to their embeddings. 

As pointed out before, the contextual information is what defines the meaning of a word. The ability to correlate words through their embeddings is especially crucial for the task of medical concept normalization. The task is concerned with finding different terms that relate to the same concept. Since they belong to the same concept they should be close in the embedding space, if the embedding model is working correctly. To find the closest concept for a term, the closest embeddings of a concept can be searched. The most common measure of similarity in embedding space is the cosine similarity \citep{singhal_modern_2001}.

The cosine similarity is a measure to quantify the relationship between two vectors and it is frequently used to compare word embeddings in the semantic space. It measures the cosine of the angle between two vectors, giving an indication of how closely related the embeddings of two words are. The cosine similarity score ranges between -1 and 1. A score of 1 implies that the vectors are identical, 0 suggests orthogonality or no correlation, and -1 means they are diametrically opposed.

To illustrate, consider two vectors $v$ and $w$ representing the embeddings of the words ``king'' and ``queen'', respectively. Since both words are semantically related, their vector representations would typically be close in the embedding space. The cosine similarity can be mathematically defined as:
\begin{equation}
\label{eq:cosine}
\text{cosine\_similarity}(v, w) = \frac{v \cdot w}{{||v||}^2 \cdot {||w||}^2} =  \frac{ \sum_{i=1}^{n}{{ v}_i{w}_i} }{ \sqrt{\sum_{i=1}^{n}{{v}_i^2}} \sqrt{\sum_{i=1}^{n}{{ w}_i^2}} }
\end{equation}
where $||.||^2$ denotes the euclidean distance and $v_i$ and $w_i$ are the \textit{i}th components of vectors $v$ and $w$, respectively.

The cosine similarity of $v$ and $w$ should be close to 1, indicating that ``king'' and ``queen'' are contextually similar in the embedding space.
 
\section{Tokenization}
\label{section:tokenization}
Next, I will attend to the process of tokenization, which on the one hand defines how an input text is split, and on the other hand constructs the vocabulary that is used to map tokens to IDs. In the last section, the example sentence \textit{I am happy} was split by whitespace into tokens and the vocabulary consisted only of four words [I, am, happy, sad]. The problem with that approach is that no words can be added to the vocabulary post hoc. This implies that the model that uses this vocabulary can not handle unseen words, which is a great limitation when applying the model to new data. One possible circumvention would be to include a very large number of words, say from a lexicon, into the vocabulary. That causes a new problem, that is well know in context of deep learning methods: \textit{Vanishing Gradients} \citep{hochreiter_long_1997}. Because an embedding model learns a probability distribution over all possible output tokens, the gradients of the loss function can become very small or zero during training. If this happens, the training signal is lost and weights are not updated correctly. 

This issue can be solved by tokens which are made-up of parts of words. \citet[p.19]{manning_introduction_2009} describe tokens as ``sequences of characters in some particular
document that are grouped together as a useful semantic unit for processing''. Tokenization is the process of splitting a text into smaller chunks of characters. These chunks of characters are mapped to IDs using the vocabulary.
 
There exist many different methods to create the tokenization rules and the vocabulary. In the following I will describe the two that are used in this thesis, namely Byte-Pair Encoding and WordPiece. 

 \subsection{Byte-Pair Encoding}
\label{section:byte_pair_encoding}

Byte-Pair Encoding (BPE) is a subword tokenization method introduced by \citet{sennrich_neural_2016}. The key idea of BPE is to create a vocabulary of subword units, starting from individual characters and iteratively merging the most frequently occurring pair of consecutive units. The merging process continues until a pre-specified number of merges is reached, which is a hyperparameter set by the user. This method is advantageous as it allows for a flexible vocabulary that can handle unseen words by breaking them down into known subword units.

BPE starts by representing each word as a sequence of characters. A special end-of-word symbol, typically ``<\textbackslash w>'', is added to each word to denote the word boundaries. The frequency of each symbol, initially only individual characters and the end-of-word symbol, is recorded in a dataset. The algorithm then iteratively merges the most frequent pair of consecutive symbols to form a new symbol, and updates the symbol frequencies accordingly. This process is repeated until the desired vocabulary size is reached.

The output is a list of merges, from which a subword vocabulary can be constructed. When tokenizing new text, BPE finds the longest subwords that appear in the merge list, starting from the beginning of the word. If no subword is found, the algorithm falls back to the individual characters.

The advantage of BPE is that it can handle out-of-vocabulary words effectively. Since it breaks down words into smaller subword units, it can represent unknown words by a sequence of known subword units. This makes BPE robust against typos and variations in word forms. However, BPE does not take into account any semantic information when constructing the subword units. Subword units often contain linguistic information, e.g., morphemes. An example would be an unknown word with the ending \textit{-ing}. Although the root of the word might not be in the vocabulary, there is a high chance that \textit{-ing} was added to the vocabulary.

\subsection{WordPiece}
\label{section:wordpiece}

WordPiece by \citet{wu_googles_2016} is a subword tokenization method that became popular by its use with the BERT architecture. It is an extension of the Byte-Pair Encoding (BPE) method, with a notable difference in the way it selects subwords for the vocabulary.

Similar to BPE, WordPiece begins with a base vocabulary of individual characters and iteratively merges the most frequent pairs of consecutive tokens. However, the criterion for the selection of pairs to merge is distinct. Instead of simply choosing the most frequent pair, WordPiece aims to choose the merge that maximizes the likelihood of the training data given the current vocabulary.

\begin{figure}[h]
  \centering
  \includegraphics[width=\linewidth]{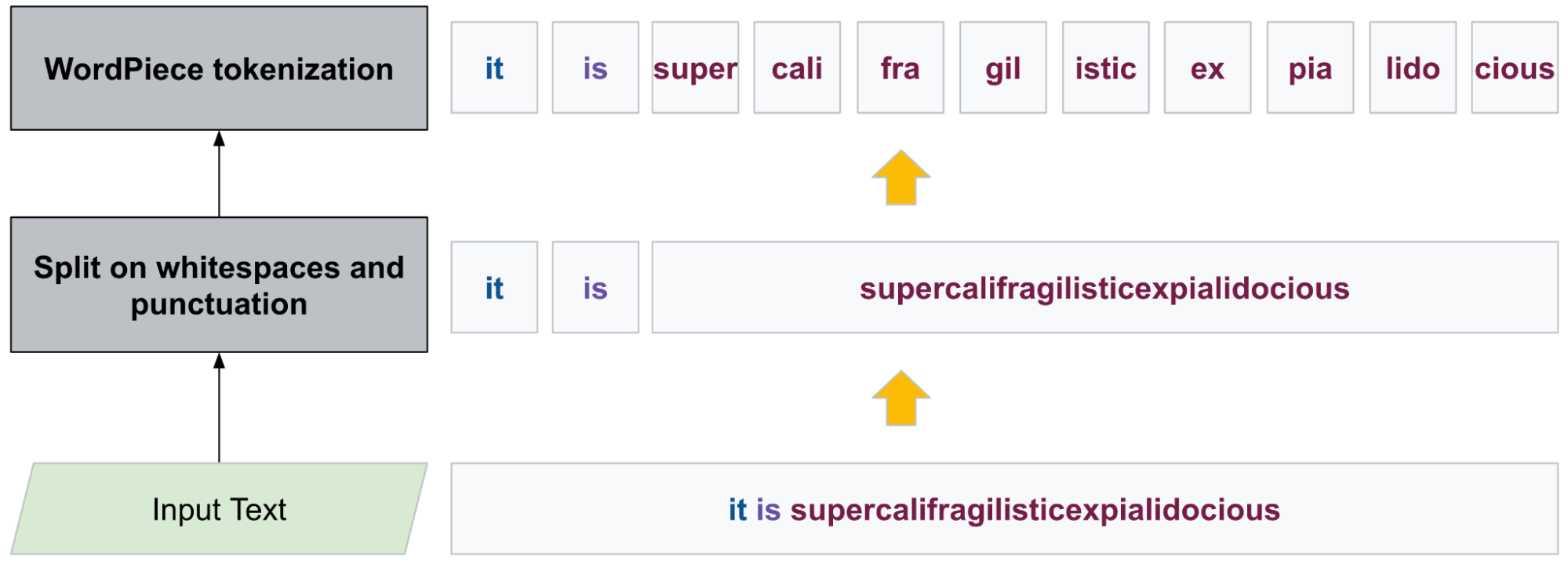}
  \caption[WordPiece tokenization]{The WordPiece tokenization process with an example sentence. Taken from a \href{https://ai.googleblog.com/2021/12/a-fast-wordpiece-tokenization-system.html}{Google blogpost}.}
\label{fig:wordpiece}
\end{figure}

The original WordPiece algorithm starts by defining a large base vocabulary, e.g., all unigrams in the training data, and then learns a fixed-size vocabulary on top of this base vocabulary that maximizes the likelihood of the training data. It does so by repeatedly choosing the best token pair to merge, where the "best" pair is the one that, when merged, causes the smallest decrease in the likelihood of the training data.

The WordPiece method addresses some of the limitations of BPE by considering the impact of a merge operation on the overall likelihood of the training data. This makes the algorithm more sensitive to the semantics of the language, which could result in more meaningful subword units.

One challenge with WordPiece is that it requires a large amount of computational resources to determine the best pair to merge. This is because the likelihood of the training data needs to be recalculated for each potential merge operation. As BPE, it is effective at handling of out-of-vocabulary words and can also be used to tokenize text in different languages \citep{wu_googles_2016}.

\section{Transformers}
The Transformer architecture, introduced by \citet{vaswani_attention_2017} \footnote{The entire section is based on this work.}, is a novel neural network architecture in NLP that has demonstrated significant performance improvements and decreased training times compared to prior methods like Recurrent Neural Networks \citep{hochreiter_long_1997} and Convolutional Neural Networks \citep{fukushima_neocognitron_1980}. 

The Transformer architecture consists of two main components: the encoder and the decoder. Input data is pre-processed by tokenization and a positional encoding, defined by a pair of sine and cosine functions, is added to the embeddings to provide information about the order of the sequence.

\begin{figure}[ht]
  \centering
  \includegraphics[width=140pt]{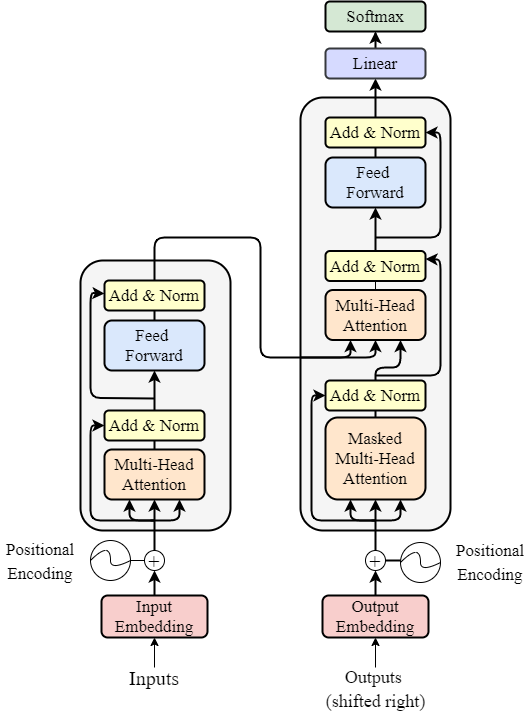}
  \caption[Transformer Architecture]{The Transformer architecture. Taken from \citet{vaswani_attention_2017}.}
\label{fig:transformer_architecture}
\end{figure}

The encoder is composed of a stack of identical layers, each consisting of a Multi-Head Attention\footnote{For an explanation of the Multi-Head Attention I refer to the original paper of \citet{vaswani_attention_2017}.} layer followed by a fully-connected neural network. Each sub-layer includes a residual connection for additional information flow, and their outputs are normalized before being passed to the subsequent layer. The encoder maps its input to an embedding representation that contains all the learned information.

The decoder, also made up of identical stacked layers with Multi-Head Attention, generates text sequences. The decoder, as the name suggests, decodes the embedding to text input. In this work, I am only interested in the embeddings produces by Transformers, so the decoder will not be further explained.

The original work by \citet{vaswani_attention_2017} demonstrated that attention-based models outperform previous NLP models and train faster than other types of neural networks, establishing the Transformer as a dominant paradigm in recent NLP.

\subsection{BERT}
\label{section:BERT}

Bidirectional Encoder Representations from Transformers (BERT) is a model proposed by \citet{devlin_bert_2019}. It is pre-trained on unlabeled text, which is intended to be easily fine-tuned for specific downstream tasks. The base version of BERT consists of a stack of 12 encoders, each employing 12 Multi-Attention Heads, and outputs embeddings of dimension 768, yielding a total of approximately 110 million parameters. The large version consists of a stack of 24 encoders, each with 16 Multi-Attention Heads, has an output dimension of 1024, and amounts to 336 million parameters.

BERT is an instance of the Transformer architecture and is trained with two objectives: Masked Language Modelling (MLM) and Next Sentence Prediction (NSP). The MLM task aims to predict masked tokens in an input sequence. For instance, in the sequence "The horse raced past the barn fell", 15\% of the tokens are randomly selected and replaced with a [MASK] token. The model is trained to predict the original token that was masked. This procedure allows the model to condition on context tokens from the left and right of the predicted token.

The NSP task is beneficial for understanding the relation between two sentences, which is crucial for tasks like Question Answering. In NSP, given a pair of sentences A and B, the model is trained to determine whether they are consecutive sentences.

\begin{figure}[h]
  \centering
  \includegraphics[width=140pt]{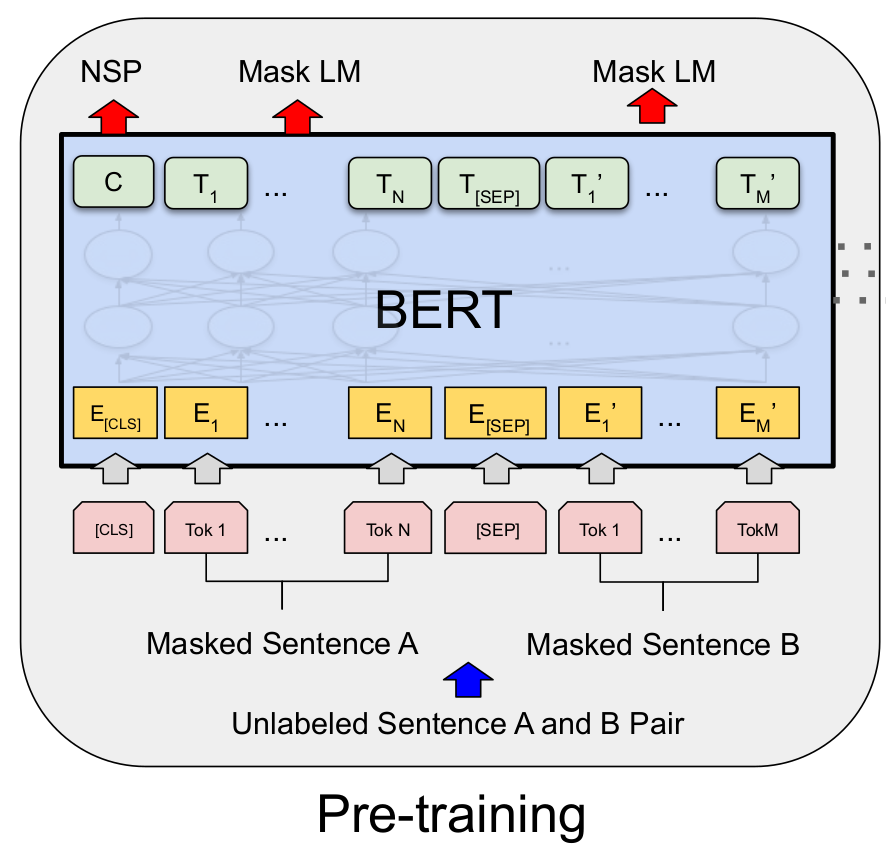}
  \caption[BERT Pre-Training]{The pre-training setup for BERT. The position of the \texttt{[CLS]} and \texttt{[SEP]}  and both training objectives are shown. Taken from \citet{devlin_bert_2019}.}
\label{fig:bert_architecture}
\end{figure}

BERT uses a separate cross-entropy loss for each task. The final loss in pre-training is a linear combination of these two losses. To accomplish these tasks, the input embeddings are modified by adding a position embedding and a segment embedding to each token embedding. Two special tokens, \texttt{[CLS]} and \texttt{[SEP]}, are added to indicate the beginning of the sequence and the end of a sentence, respectively. The final embedding of the \texttt{[CLS]} token, used for the final classification within a classification task, contains most of the information in an output sequence. The final embedding of \texttt{[CLS]} is passed to a fully-connected layer with an activation function and can be used to predict a class.

The pre-trained base and large BERT model provided by the authors were trained on the a corpus of books (800 Million words) and the English Wikipedia (2.500 Million words), which were pre-processed so that the data only contained text passages.

\subsection{Robustly Optimized BERT Approach}
\label{section:roberta}
A commonly used BERT-based architecture is Robustly Optimized BERT Approach (RoBERTa) by \citet{liu_roberta_2019}.\footnote{The entire section is based on \citet{liu_roberta_2019}} While RoBERTa shares the same architecture as BERT, it distinguishes itself through three extensions, that were intended to boost the performance over BERT.

RoBERTa benefits from training on a dataset encompassing 160GB of text, marking a significant expansion, over ten times larger than BERT's training dataset. 
Furthermore, it employs a dynamic masking technique during training. Unlike BERT, which applies a static mask during data preprocessing, RoBERTa's training data undergoes the masking process multiple time, such that in each iteration different tokens of the same sentence are masked. This should improve the efficiency of the training data and make the model more robust. 

And lastly, the Next Sentence Prediction objective is discarded in RoBERTa. The authors report that omitting NSP either retains or slightly augments performance on downstream tasks.

\subsection{Multilingual Transformers}
This short section should state how the two multilingual models I used in this thesis are trained.

Multilingual BERT (mBERT) uses the exact same architecture which is described as the base model in Section \ref{section:BERT}. It is trained on a large multilingual corpus, comprising Wikipedia data in 104 languages.\footnote{There exists no published work that further describes the training data, but a list of languages that where present in the training data can be found \href{https://github.com/google-research/bert/blob/master/multilingual.md}{in this document from Google}.}

The multilingual RoBERTa model, RoBERTa\textsubscript{XLM}, by \citet{conneau_unsupervised_2020} was trained on more than two terabytes of filtered Common Crawl \citep{wenzek_ccnet_2020} data. The Common Crawl dataset is a multilingual and diverse dataset that includes petabytes of scraped web pages, ranging from academic papers and social media texts to coding examples. The training data consisted of 100 different languages. The authors report an accuracy gain of 2.4\% to 14.6\% on natural language inference, question answering and named entity recognition tasks over mBERT.

\section{Refining Word Embeddings}
\label{section:self_alginment_pretraining}
The former three sections explained how powerful models are trained that create general-purpose word embeddings. They set new benchmarks across many different tasks in NLP \citep{devlin_bert_2019, phan_robust_2019}. Applying them to biomedical texts might yield inferior results, because they are trained on common language in an self-supervised setting. There is no mechanism that ensures that they capture the fine-grained semantic relationship of biomedical entities during pre-training. 

\begin{figure}[h]
  \centering
  \includegraphics[width=0.8\linewidth]{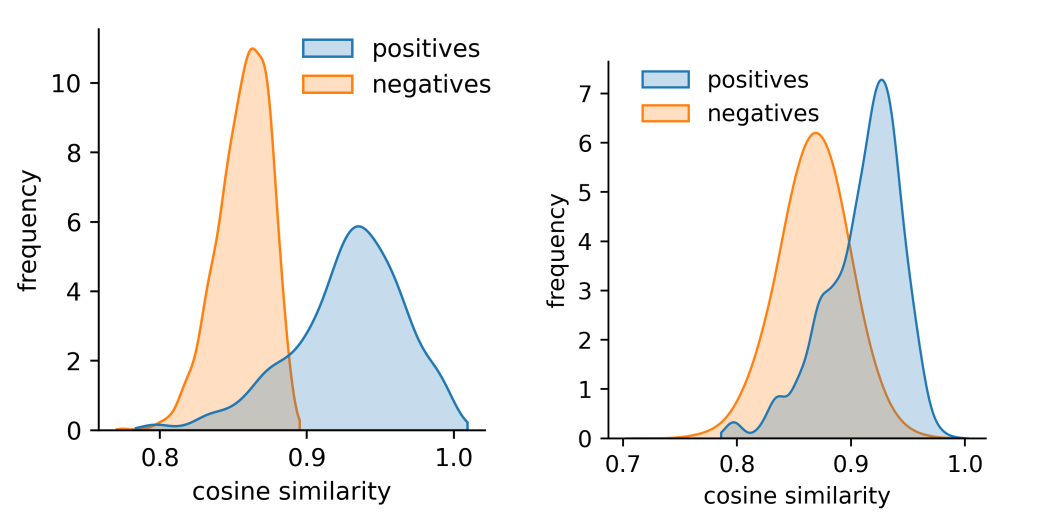}
  \caption[SapBERT Embedding Visualization]{Cosine similarity scores of pairs of embeddings produced by PubMedBERT. The examples were randomly samples from UMLS. Positive pairs consist of names that belong to the same concept and negative pairs consist of names from two different concepts. The left graph shows examples of embeddings that are already well-separated in the embedding space. The right diagram shows examples with a high overlap between positive and negative samples. These are the \textit{hard} examples which are selected by the hard pairs mining sampling. Taken from \citet{liu_self-alignment_2021}.}
\label{fig:mining_examples_overlap}
\end{figure}

A desirable property of a embedding model for MCN is that it creates similar embeddings for the name and synonyms of the same concept in the knowledge base. At the same time, embeddings for names and synonyms that do not belong to the same concept should be well-separated. Figure \ref{fig:mining_examples_overlap} shows the distribution of similarity scores for pairs of samples that are either well-separated (left graph) or show a high overlap of similarity scores (right graph). Sample pairs with a high overlap are prone to be assigned to the wrong class because the embedding of wrong concept might be closer to the sample than the embedding of the correct class.

There exist methods that try to accomplish exactly this separation \citep{liu_self-alignment_2021, liu_learning_2021, yuan_coder_2022}. In my work, I will extensively work with the \textit{Self-Alignment Pre-training} method, proposed by \citet{liu_self-alignment_2021} and extended by \citet{liu_learning_2021}, which will be explained in the following.\footnote{The entire section is based on the work of \citet{liu_self-alignment_2021}.}

The \textit{Self-Alignment Pre-training} is a fast and effective pre-training method designed to align synonymous biomedical entities from a knowledge base. Starting with an pre-trained embedding model like BERT, it learns a function that maps a name to a embedding space. The weights of the embedding model are changed such that synonyms are close in the embedding space. The synonyms are taken from the knowledge base, e.g. UMLS, that is used to pre-train the embedding model.

The \textit{Self-Alignment Pre-training} method relies on two building blocks: online hard pairs mining and the multi-similarity loss. Given a pair of tuples $\langle (x_i, y_i), (x_j, y_j)\rangle$, the similarity of $x_i, x_j$ can be measured by the cosine similarity (Equation \ref{eq:cosine}) 
\newline
$cosine\_similarity(x_i, x_j)$. It should be high if $y_i = y_j$ and low if $x_i \neq x_j$. 

The online hard pairs mining efficiently selects informative training examples, i.e., most hard negative or positive sample pairs. The samples are selected as triplets $x_a, x_p, x_n$, where $x_a$ is called an anchor sample and $x_a$ and $x_n$ are a hard positive and negative sample, respectively. The sampling of the triplets is constrained by the condition 
\begin{equation}
\label{eq:similarity_constrait}
{||f(x_a) - f(x_p)||}^2 < {||f(x_a) - f(x_n||}^2 + \lambda
\end{equation}
Triplets are sampled such that the negative sample $x_n$ is closer to the anchor $x_a$ than the positive sample $x_a$ by a margin of $\lambda$. Each triplets yields two training example pairs. The set of hard positive pairs $x_a, x_p$ is denoted as $\mathcal{P}$ and the set of hard negative pairs $x_a, x_n$ as $\mathcal{N}$. In an ablation study, the authors compare how the online hard pairs mining compares to random sampling. \citet{liu_self-alignment_2021} use PubMedBERT as embedding model and apply \textit{Self-Alignment Pre-training} with and without their sampling procedure on the UMLS knowledge base. Afterwards they compare both models on the test set of the COMETA dataset. They report an accuracy gain of 14.9 absolute accuracy points using their sampling procedure over random sampling. 

To adapt the weights of the embedding model with respect to the hard and positive sample pairs, the multi-similarity loss (MS loss) by \citet{wang_multi-similarity_2019} is used. First a similarity matrix of the embeddings of all concept names in the knowledge base is created. In practice, the similarity matrix is calculated on one batch of concept names. The similarity matrix is denoted as $S \in \Re^{M \times M}$ where M is the number of names in the knowledge base and each entry $S_{ij}$ corresponds to the similarity between the \textit{i}-th and \textit{j}-th concept name or synonyms. The MS loss adapted to include $\mathcal{N}$ and $\mathcal{P}$ is: 
\begin{equation}
\label{eq:ms_loss}
    \mathcal{L} = \frac{1}{M} \sum_{i=1}^{M}(\frac{1}{\alpha} log (1 + \sum_{n \in \mathcal{N_i}} e^{\alpha(S_{in}-\epsilon)}) + \frac{1}{\beta} log ( 1 + \sum_{p \in \mathcal{P_i}} e^{-\beta(S_{ip}- \epsilon)})) 
\end{equation},
where $\alpha$, $\beta$ are temperature scales, $\epsilon$ is the offset of the similarity matrix, and $\mathcal{P}_i, \mathcal{N}_i$ are the respective positive and negative samples of the anchor sample $i$.

The loss is calculated on the similarity matrix S, the gradient of the loss function with respect to the input is derived and the weights of the embedding model are updated.

\section{Sentence Cross-Encoders}
\label{section:sentence_cross-encoder}

Sentence Cross-Encoder by \citet{reimers_sentence-bert_2019} is a method to efficiently compute text similarity of two texts.
A Sentence Cross-Encoder employs a Transformer-based architecture into a siamese network structure with triplet loss \citep{schroff_facenet_2015}. To find the most similar items out of a set of texts, a common approach is pass each possible pair of text to a BERT-based model and derive the similarity score on the output representation of the \texttt{[SEP]} token. As explained in Section \ref{section:BERT}, BERT is trained with the next sentence prediction objective on sentence pairs. The NSP loss guides the model to embed information about the similarity of the two input sentence in the \texttt{[SEP]} token representation. The method is not easily applicable to concept normalization because the computational time to find the closest concept for one mention depends on the number of concepts in the knowledge base. With a knowledge base containing $n$ concepts and $m$ mentions for which we want to find the closest concept, the BERT model needs to be queried $m \cdot n$ times. This is because in this approach the embedding of each concept depends on the mention it is compared to, hence the embedding of each concept needs to be recalculated for each mention. The Sentence Cross-Encoder methods solves this problem by creating a fixed-sized embedding of the concepts that is independent from the mentions. In that way, the embeddings of the concepts can be calculated and cached beforehand. Finding the most similar concept for $m$ mentions, consists now only of embedding the $m$ mentions into the embedding space and finding the closest concept for each mention using a similarity measure like cosine similarity. 

\begin{figure}[h]
  \centering
  \includegraphics[width=0.6\linewidth]{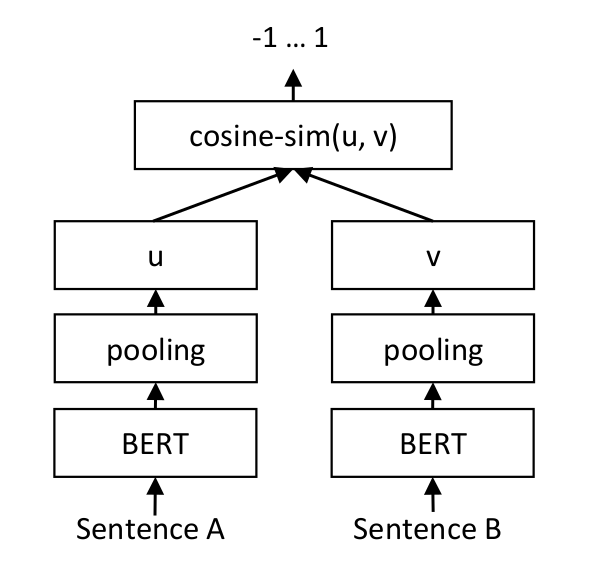}
  \caption[Sentence Cross-Encoder Architecture]{The Sentence Cross-Encoder architecture with a BERT encoder to calculate similarity scores. The weights of the two BERT model are tied (siamese network). Taken from \citet{reimers_sentence-bert_2019}.}
\label{fig:sentence_encoder_comparison}
\end{figure}

A Sentence Cross-Encoder to produce similarity scores for mention concept pairs can be trained as follows: The classification head of a vanilla BERT is replaced with a pooling layer. The work of \citet{reimers_sentence-bert_2019} contains an experiment comparing different pooling methods. In my experiments I will use mean pooling which showed the best performance in their work. The triplet loss function is directly applied to the pooled output representations of two mentions. The input pairs are selected in triplets where for each anchor sample $x_a$, a sample from a different class $x_n$ and from the same class $x_p$. The loss function to minimize is defined by the triplet loss on the input triplet:
\begin{equation}
     loss = max({||f(x_a) - f(x_p)||}^2 < {||f(x_a) - f(x_n||}^2 + \lambda,0)
\end{equation}
where $f$ is the pooled BERT output and the other parameters are defined as in Equation \ref{eq:similarity_constrait}. $\lambda$ is typically set to 1.

\section{Evaluation Metrics for MCN}

In this section, the metrics used to evaluate the performance of a model for MCN are presented. As outlined in Section \ref{section:related_mcn_methods}, there exist different approaches to MCN. However, the final assignment of one concept to a mention can always be evaluated by a multi-class classification metric. Here, I will use \textit{Accuracy}, \textit{Accuracy@n} and \textit{F\textsubscript{1}-score} as the classification metrics. \citet{powers_evaluation_2008} define them as:

\begin{equation}
Accuracy = \frac{\sum_{i=1}^{C} TP_i}{\sum_{i=1}^{C} (TP_i + FP_i + FN_i)}
\end{equation}

where $C$ denotes the number of classes, and $TP_i$, $FP_i$, and $FN_i$ are the number of true positives, false positives, and false negatives, respectively, for class $i$.

\textit{F\textsubscript{1}-score} is the harmonic mean of Precision and Recall. It is calculated as:

\begin{equation}
F\textsubscript{1}-score = 2 \cdot \frac{Precision \cdot Recall}{Precision + Recall}
\end{equation}

where

\begin{equation}
Precision = \frac{\sum_{i=1}^{C} TP_i}{\sum_{i=1}^{C} (TP_i + FP_i)}
\end{equation}

and

\begin{equation}
Recall = \frac{\sum_{i=1}^{C} TP_i}{\sum_{i=1}^{C} (TP_i + FN_i)}
\end{equation}

The Accuracy@n, or Top-N Accuracy, is another metric that is often used within medical concept normalization. This metric assesses whether the correct concept is within the top N predicted concepts. It is only applicable to models that output an ordered set of concepts, which is the case with all the models used in this thesis.

Furthermore, concept normalization is a multi-class classification problem. The metrics Precision, Recall and F\textsubscript{1}-score are calculated with respect to one class, typically in an one-vs-rest manner. Hence one has to decide how to average the metrics over all classes. 
Being robust to class imbalance is a desirable property when normalizing concepts to a knowledge base, because often the number of instances is much smaller than the number of concepts. Concepts that appear multiple times in a dataset can quickly lead to imbalanced distribution of classes. I decided to use \textit{weighted} averaging which aggregates the metric for each class and weights them by the number of samples that were labeled with that class. The weighted F\textsubscript{1}-score is calculated as a weighted sum of the individual F\textsubscript{1}-scores for each class.

\chapter{Data Collection}
\label{section:data_collection}

\section{Annotating TLC with UMLS Concepts}
In this section, I will outline the process of data collection and annotation. The data to annotate is the Technical-Laymen Corpus (TLC), an annotated forum dataset sourced from \href{Med1.de}{Med1.de}, a German online health platform where non-professional users exchange advice and opinions on a range of health topics. The TLC focuses on medical language used by patients and medical laymen, with a particular emphasis on two subforums dedicated to kidney diseases and stomach and intestines.

These subforums comprise user questions and corresponding answers, providing a rich source of both technical and lay medical language. This data can give valuable insights into the interplay between professional and lay medical terminology, offering a realistic view of how medical language is used in non-professional contexts. The forum data include the time of posting, author's username, and thread title, with potential identifiers anonymized to protect user privacy. In total, TLC contains 4,000 forum posts, containing 6,472 annotations, with an average of 1.62 annotations per post. The annotations contains two labels: lay medical terms and technical medical terms. In total there are 4.170 lay annotations and 2.302 technical annotations. Additionally, one or more synonym is provided for each mention. If the mention is in lay language, most of the time it is the technical version of the mention. If the mention is a technical language, a technical or rarely a lay alternative name is provided. One random sample from the dataset is the following text: 
\begin{quote}
    Nierenkrank mit 26? Ich schlafe nicht ich esse nicht und ich geh nicht nach draußen vor lauter Panik. Ganz ehrlich? Ich denke, dein Kreatininwert ist im Vergleich zu deiner Panik ein sehr, sehr kleines Problem.
\end{quote}
\begin{quote}
    \textbf{English Translation:} Kidney disease at 26? I don't sleep, I don't eat, and I don't go outside for panic. Honestly? I think your creatinine is a very, very small issue compared to your panic.
\end{quote}
In this sample, the first word "Nierenkrank" is annotated as lay mention with the additional synonym "Niereninsuffizienz" (renal failure).

However, for the task of medical concept normalization, it is necessary to further annotate mentions with unique concept identifiers (CUIs) from UMLS. The following sections detail the methodology and results of this annotations process.

\subsection{Preparing TLC for UMLS annotations}
There are duplicate mentions in the dataset that do not need to get annotated twice with the correct UMLS concept by the annotators. But the meaning of a mention can differ in a different context. Hence, it is not trivial to find the correct duplicates.  
To decide whether two mentions are equal, not only the stemmed mentions are compared, but also the stemmed synonyms with which they are annotated. If a mention is technical, it is compared to other technical mentions or their technical synonyms. If a mention is lay language, only its synonyms are compared. By doing so, I created a set of 997 unique mentions to be annotated. If the procedure is repeated with stemmed mentions and synonyms, the final set of unique mentions consists of 851 items.

To pre-annotate mentions with UMLS concepts, the following procedure was adopted: each unique mention and its synonyms from the TLC dataset were first translated into English. The UMLS database was then queried using its official API\footnote{\href{https://uts.nlm.nih.gov/uts/umls}{https://uts.nlm.nih.gov/uts/umls}}. The priority for matches was as follows: original mention \textgreater \ original synonym \textgreater \  translated mention \textgreater~translated synonym. In cases where no matches were found, a local Solr system storing all German UMLS concepts was used. The original synonym was queried, and \textit{Edit Distance} served as the similarity measure to ensure a concept match, even if the similarity score was low.

Ultimately, each unique mention received a pre-annotation consisting of a UMLS Concept Unique Identifier (CUI). A two-step process was then carried out for each pre-annotated CUI. First, the concept description was sourced from available German vocabulary; if unavailable, a DeepL\footnote{https://www.deepl.com/docs-api} translation of the UMLS description was used. Then, the UMLS preferred name was located and translated into German. This workflow yielded each pre-annotation with the unique mention, original text, a CUI, a German preferred term, and a German description.

\subsection{Annotation Setup}
\begin{figure}[h]
  \centering
  \includegraphics[width=\linewidth]{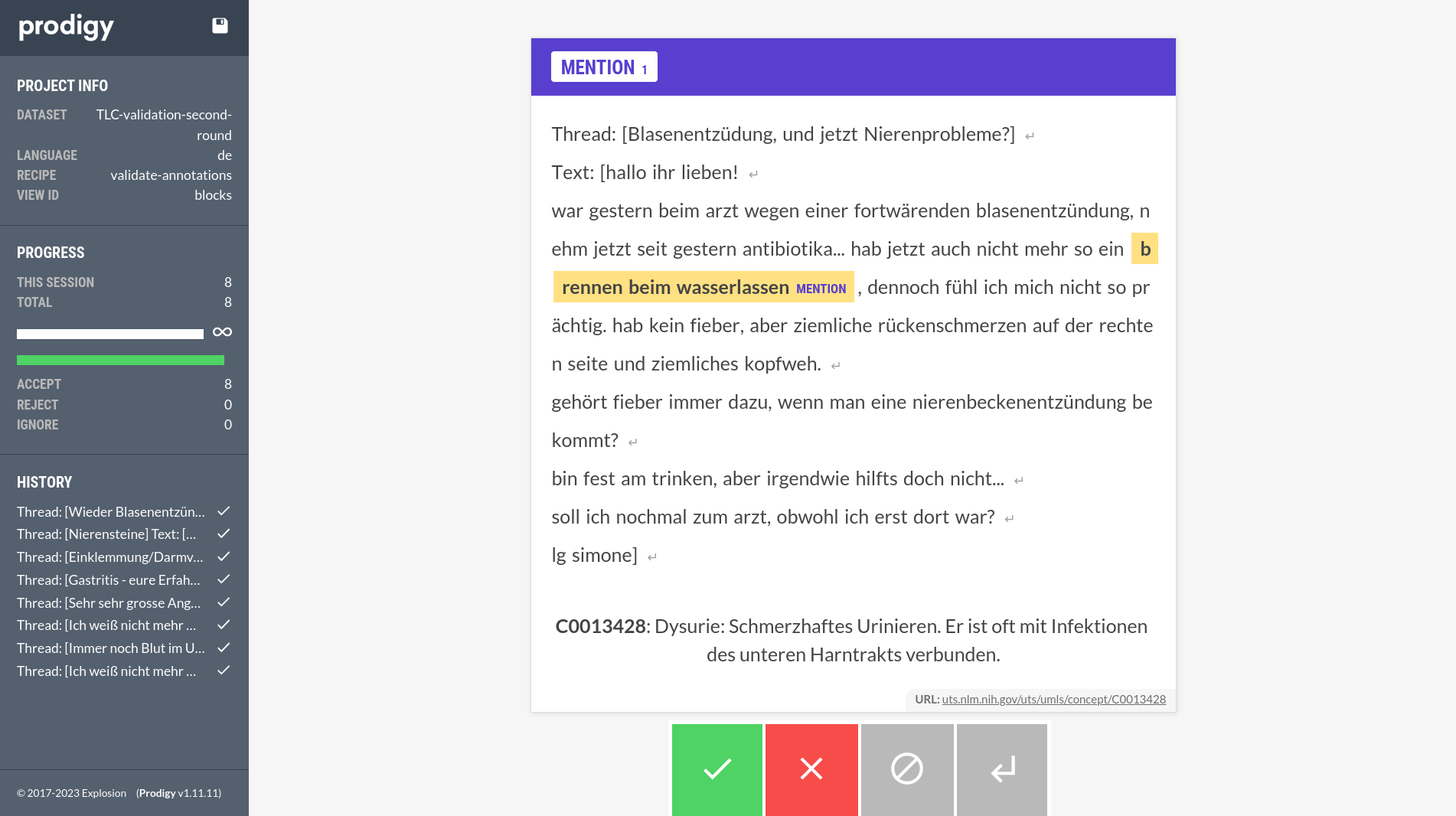}
  \caption[Prodigy Annotation Interface]{The annotations interface by Prodigy as it is presented to the annotators. The annotators are expected to accept or reject an annotation using the buttons on the bottom of the page.}
\label{fig:prodigy_interface}
\end{figure}

As annotation tool Prodigy v1.11.0 \citep{montani_prodigy_nodate} is used. Prodigy allows users to annotate text, images, and other data formats to create labeled datasets. It provides an intuitive and interactive interface that simplifies the annotation process, making it easier for humans to label data accurately and efficiently. The tool supports a wide range of annotation tasks, including named entity recognition, text classification, sentiment analysis, and more. A view of the annotation interface that is presented to the annotators is shown in Figure \ref{fig:prodigy_interface}.
The annotation task is to either accept or reject each suggested annotation with the respective span in the post. The annotators are presented with the entire post in which the mention is marked. Shown below the post, there is the suggested CUI along with its description from UMLS. The spans of the mention could not be changed by the annotators. Each annotator was presented with all 851 mentions.

After the first validation round by the annotators, 523 annotations were mutually accepted and were marked as ready to go directly into the new TLC-UMLS dataset. All others need to be refined in a second annotation step. There were 105 annotations which were rejected by either one of the annotators. These were discussed by both annotators and then again either accepted or rejected. The accepted ones were marked as ready. For the new rejected ones and the ones for which both annotators agreed on reject in the first round, I created new annotations manually. Here manually means querying the UMLS database by hand for suitable concepts. The new annotations were again presented to the annotators and they were asked to either accept or reject them. All annotations except for ten were accepted by both this time. The accepted ones were marked as ready and the remaining ones were manually annotated by the annotators. 

\subsection{Evaluation of The Annotation Process}

Analyzing the agreement of the annotators, Annotator 1 and Annotator 2 had a substantial agreement on 623 instances where both accepted, and 94 instances where both rejected. However, there were 64 instances accepted by Annotator 1 and rejected by Annotator 2, and 51 instances accepted by Annotator 2 and rejected by Annotator 1. The number of individual accepted and rejected annotations were very similar. Annotator 1 accepted 687 and rejected 145 annotations in total, while Annotator 2 accepted 874 and rejected 138 annotations. A confusion matrix depicting how many mentions were accepted and rejected by each annotator can be found in Figure \ref{table:annotator_confusion_matrix}

F\textsubscript{1}-score and Cohen's Kappa \citep{cohen_coefficient_1960} are used for evaluating the inter-annotator agreement in normalizing tasks. Cohen's Kappa was \textbf{0.67} and average F\textsubscript{1}-score \textbf{0.9} for accept and \textbf{0.77} for reject. The F\textsubscript{1}-score of accept and reject would be the other way round if Annotator 1 instead of Annotator 2 would be used as gold standard. According to \citet{mchugh_interrater_2012}, a Cohen's Kappa of 0.67 indicates a substantial level of agreement between the two annotators. The kappa statistic adjusts for the amount of agreement that could be expected to occur through chance, providing a more robust measure than simple percent agreement. In the context of inter-annotator reliability, a kappa of 0.67 suggests that the annotators largely agreed on the classification of mentions. However, there is still some room for improvement, as perfect agreement would be indicated by a kappa of 1. The F\textsubscript{1} score, on the other hand, provides a performance measure for each individual annotator by considering both precision (how many selected items are relevant) and recall (how many relevant items are selected). The average F\textsubscript{1} score of 0.86 suggests that the annotators were generally accurate in their classification of mentions as \textit{accept} or \textit{reject}. This high F\textsubscript{1} score indicates that both the precision and recall were high, meaning that the annotators were able to correctly identify most of the relevant mentions and were generally correct when they identified a mention as relevant. Again, an F\textsubscript{1} score of 1 would indicate perfect precision and recall, so while 0.86 is relatively high, there is still potential for improvement.

\begin{table}
\caption[Annotation Confusion Matrix]{Confusion Matrix of how many mentions were accepted or rejected by each annotator.}
\label{table:annotator_confusion_matrix}
\centering
\begin{tabular}{l|l|c|c|c}
\multicolumn{2}{c}{}&\multicolumn{2}{c}{Annotator 1}&\\
\cline{3-4}
\multicolumn{2}{c|}{}&Accepted&Rejected&\multicolumn{1}{c}{Total}\\
\cline{2-4}
\multirow{2}{*}{Annotator 2}& Accepted & $623$ & $51$ & $674$\\
\cline{2-4}
& Rejected & $64$ & $94$ & $138$\\
\cline{2-4}
\multicolumn{1}{c}{} & \multicolumn{1}{c}{Total} & \multicolumn{1}{c}{$687$} & \multicolumn{    1}{c}{$145$} & \multicolumn{1}{c}{$822$}\\
\end{tabular}
\end{table}

\section{The TLC-UMLS dataset}
The new dataset will be called \textbf{TLC-UMLS}. It contains 6.428 annotations in 3.009 posts. Note that the original TLC dataset contained posts that did not contain any annotations. These were not added to TLC-UMLS, hence it contains less, 3.009 instead of 4.000, posts than TLC. The dataset contains 4.727 annotations that are tagged as \textit{lay} and 1.701 annotations that are tagged as \textit{technical}. The 4.727 lay mentions are 602 unique mentions while the 1.701 technical mentions are 220 unique ones.

\begin{figure}[h]
  \centering
  \includegraphics[width=0.7\linewidth]{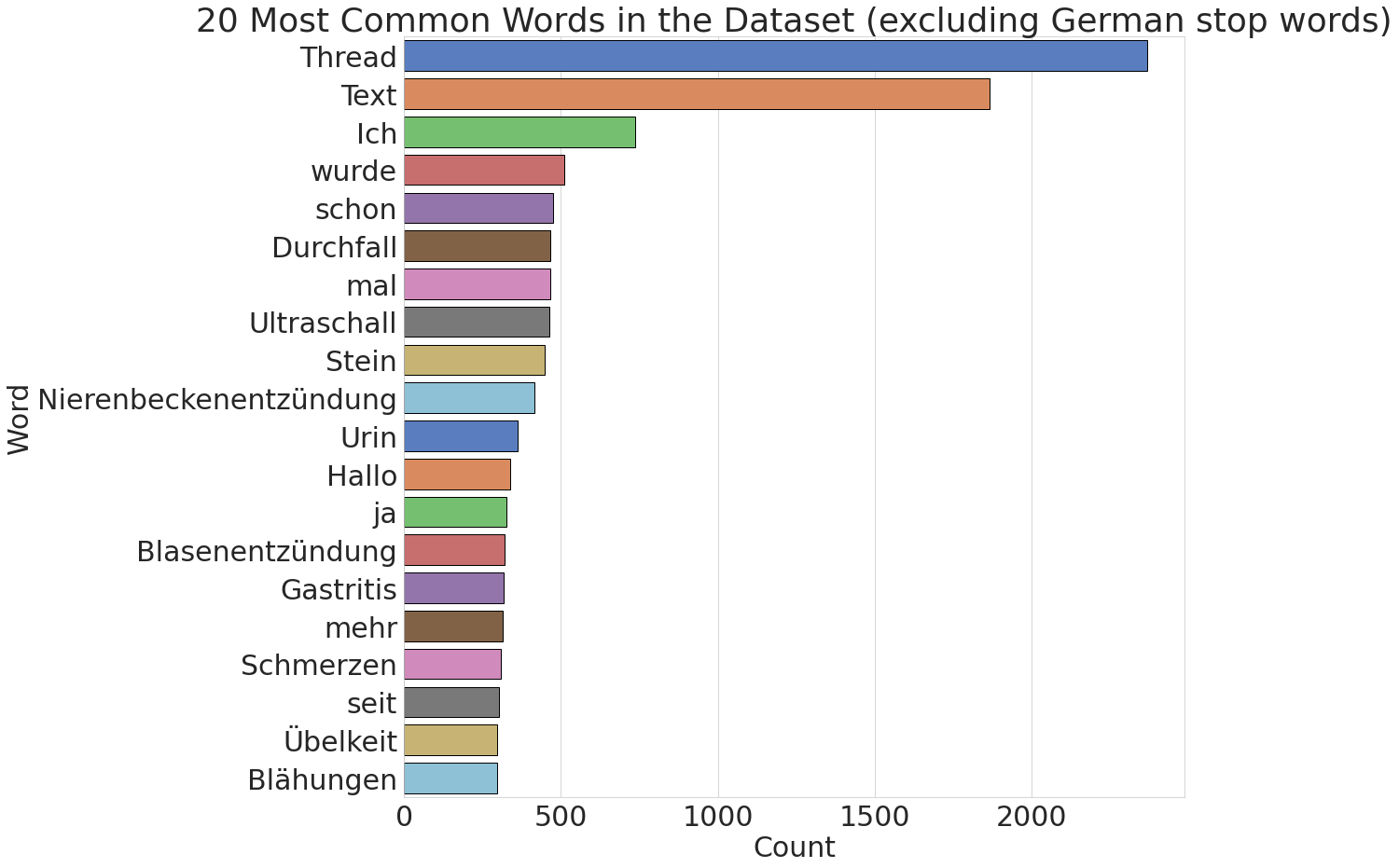}
  \caption[Most Common Words in TLC-UMLS]{The 20 most common words present in the TLC-UMLS dataset.}
\label{fig:most_common_words}
\end{figure}

In the following I will present a brief analysis of the words and concept labels that constitute TLC-UMLS. \footnote{The two most common words are ``Thread'' and ``Text'', which appear 2370 and 1868 times, respectively. This is no surprise because every post contains two sections that are prefaced with exactly these two words. Some posts are only constituted of the name of the thread in which they appear so that ``Thread'' appears more often.} As the dataset is derived from a sub-forum specialized in kidney, stomach, and intestine issues, it also unsurprisingly has a high frequency of terms related to these medical areas. Also present in the 20 most common words are the words ``ich'', ``wurde'', ``schon'' and ``mal''. These are typical words that the users of the forum use to describe their past treatments. Among the 20 most common words are also 8 words that are mentions of medical terms indicating that the content of TLC is highly related to medical subjects. The 20 most common words and their counts are shown in Figure \ref{fig:most_common_words}.

Next, I present the distribution of most common mentions and concept unique identifiers that were added to the TLC dataset.
The most common mention is ``Ultraschall'' (ultrasonography), appearing 368 times. This suggests that ultrasound, a commonly used diagnostic tool in gastrointestinal medicine, is a frequent topic of discussion on the forum.
The second most common term, ``Nierenbeckenentzündung'' (kidney infection or pyelonephritis), which appears 362 times, directly corresponds to the forum's kidney-health focus. Similarly, the term ``Stein'' (stone), appearing 354 times, likely refers to kidney stones, a common renal health issue. ``Durchfall'' (diarrhea) is another term appearing 351 times, underlining the dataset's focus on intestinal issues. ``Wassereinlagerungen'' (water retention or edema), appearing 284 times, can be related to various health conditions, including those affecting the kidneys and digestive system.

\begin{figure}[h]
  \centering
  \includegraphics[width=\linewidth]{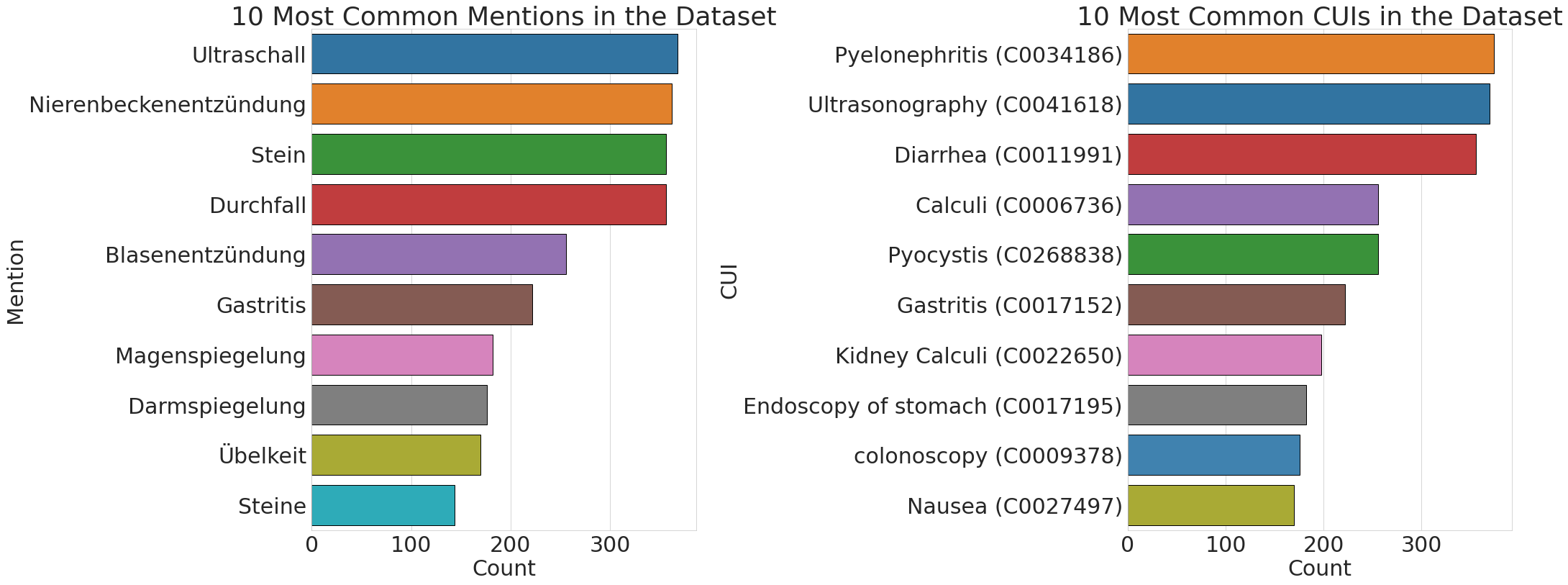}
  \caption[Most Common Mentions and Concepts in TLC-UMLS]{The 20 most common annotated mentions and the 20 most common concept that were annotated in the TLC-UMLS dataset. The color of the bar in the left chart corresponds to the color of the concepts on the right side. For example, the most common mention ``Ultraschall'' is assigned the second most common concept ``Ultrasonography''.}
\label{fig:most_common_mentions}
\end{figure}

The most common CUI in the dataset, appearing 374 times, corresponds to the medical concept of ``Pyelonephritis''. This aligns with the high frequency of the term ``Nierenbeckenentzündung'' in the dataset. The concept of Ultrasonography, represented by the second most common CUI in the dataset, appears 370 times. Figure \ref{fig:most_common_mentions} shows the 10 most common mentions along with the 10 most common concept annotated.

\begin{figure}[h]
  \centering
  \includegraphics[width=\linewidth]{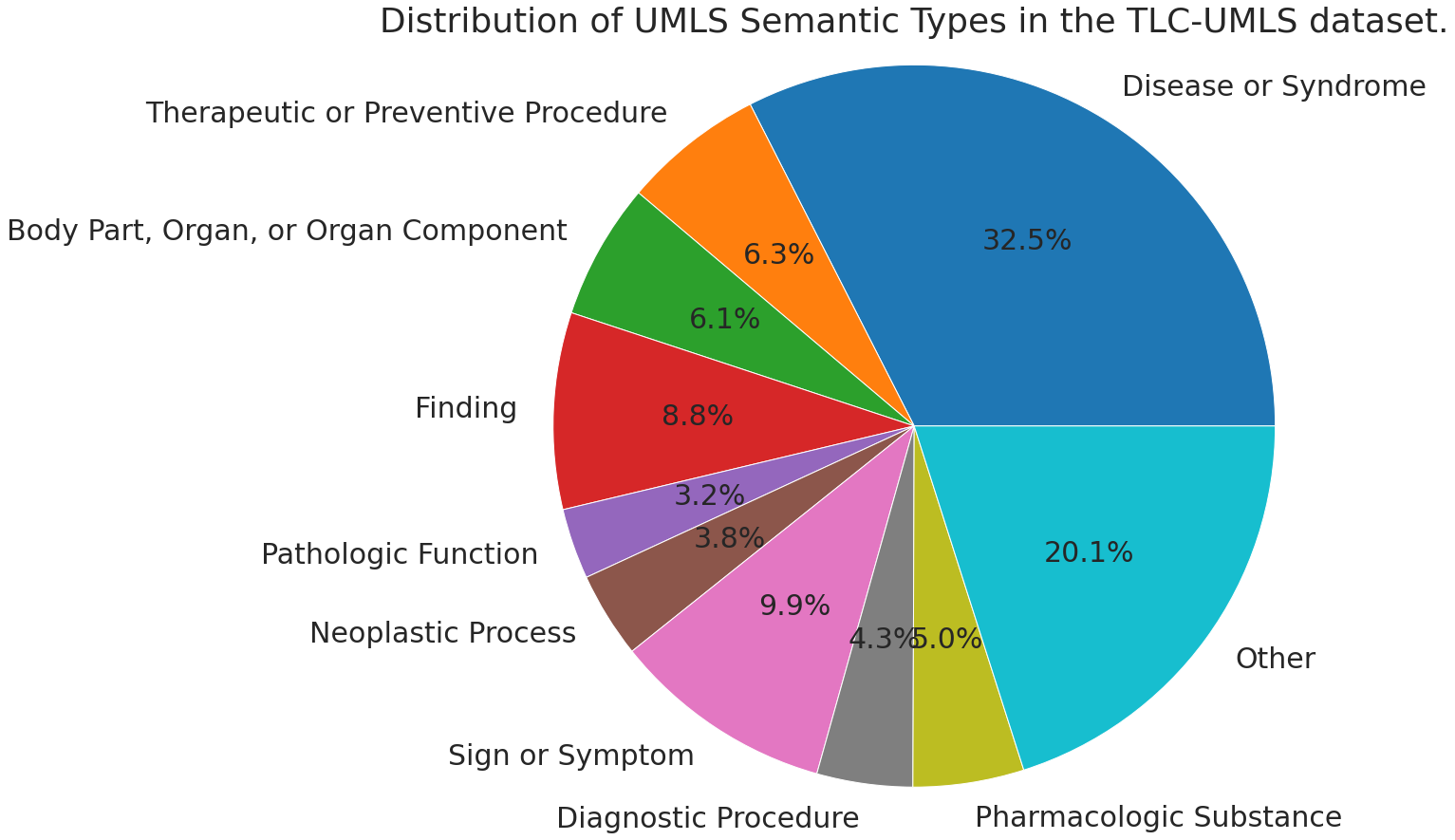}
  \caption[UMLS Semantic Type Distribution]{The distribution of UMLS Semantic Types in the TLC-UMLS annotations. In total there are 45 different types present.}
\label{fig:semantic_types_tlc_umls}
\end{figure}

Interestingly, the most common concept does not contain the most common mention, i.e. while the concept of ``Pyelonephritis'' has the highest count, the most frequent mention belongs to the concept of ``Ulrasonography''. This indicates that there exist other mentions in the dataset that belong to the most common concept. Since the TLC-UMLS dataset is supposed to be used to evaluate how concept normalization models deal with lay language, it should contain a large portion of mentions that are considered as lay. Looking at the most common mentions, ``Ultraschall'', ``Stein'' and ``Durchfall'' can be considered as terms that a medical professional would not use. 
Other common concepts in the dataset include ``Pyocystis'' or bladder inflammation, ``Gastritis'', ``Renal Insufficiency'', ``Kidney Calculi'', and ``Endoscopy of stomach''. 

It is not only interesting to look at the concepts itself but also to which semantic types they belong. As described in the section about the UMLS knowledge base, each concept is assigned to one of 127 semantic types. In TLC-UMLS are 45 different semantic types present. The share of all semantic types with more than 10 concepts in the dataset are shown in Figure \ref{fig:semantic_types_tlc_umls}. The most common semantic type is ``Disease or Syndrome'', which occurs 288 times. This is followed by ``Therapeutic or Preventive Procedure'' and ``Body Part, Organ, or Organ Component'', with 56 and 54 instances respectively. Other notably present semantic types include ``Finding'' and ``Sign or Symptom''. Some semantic types such as ``Indicator, Reagent, or Diagnostic Aid'', ``Congenital Abnormality'', and ``Bacterium'' among others are scarcely represented, each appearing only once in the dataset. The TLC-UMLS annotations represent only small subset of UMLS. Disease or syndromes as most present semantic type is a valuable property because gaining insights into diagnoses and procedures might be of more public interest than bacteria or congenital abnormalities.

\chapter{Experiments}
\label{section:experiments}

In this section I want to present how I conducted the different experiments and explain why I made certain design decisions. The experiments are designed to answer the research question ``How well do deep learning methods normalize German medical lay terms in user-generated texts linked to a large knowledge base?''. To answer that question the performance of state-of-the-art medical concept normalization models on German lay data assessed. 

In the following, I will describe how the data is pre-processed, how the baseline method is designed and how each experiment I conducted to answer the aforementioned two questions is implemented.

\section{Pre-processing TLC-UMLS and the UMLS knowledge base}

In this work I only focus on normalizing medical entities to the German subset of UMLS. The methodology can easily be extended to the entirety of UMLS concepts, but this should be the content of a follow-up research project. Downsampling UMLS has the advantage that computational demand of all methods remains in a reasonable extend, while the validity of the methods still holds true. First, all German ontologies within UMLS need to be selected and secondly, all samples from TLC-UMLS that are annotated with concepts outside of the German UMLS need to be dropped.

\textbf{Creation of a German UMLS knowledge base:} A subset of UMLS is created containing all the German source vocabularies. These are the German translation of: Medical Subject Headings (MSHGER); International Classification of Diseases, Tenth Revision (DMDICD10); Medical Dictionary for Regulatory Activities (MDRGER); WHO Adverse Drug Reaction Terminology (WHOGER); International Classification of Primary Care (ICPCGER); Universal Medical Device Nomenclature System (DMDUMD); Logical Observation Identifiers Names and Codes terminology (LNC-DE-AT, LNC-DE-DE, LNC-DE-CH).\footnote{A brief description and further information of each source vocabulary can be found under \\ \href{https://www.nlm.nih.gov/research/umls/sourcereleasedocs/index.html}{https://www.nlm.nih.gov/research/umls/sourcereleasedocs/index.html}.} The resulting subset will be called UMLS\textsubscript{DE} and contains a total of 215,401 names and 110,086 unique concepts. An exact breakdown of the share of each source vocabulary in the subset can be found in Table \ref{tab:wumls_ontologies}. 
\begin{table}[h]
\centering
\begin{tabular}{|lrrr|}
\toprule
  Ontology & Name Count & Concept Count & Share Matches (in \%) \\
\midrule
    MDRGER &      99061 &         52249 &                            31.9 \\
    MSHGER &      80864 &         39852 &                            32.9 \\
  DMDICD10 &      11864 &         11208 &                             3.81 \\
 LNC-DE-DE &      11059 &         11043 &                             0.3 \\
 LNC-DE-CH &       4941 &          4941 &                             0.4 \\
    DMDUMD &       3373 &          3296 &                             2.01 \\
    WHOGER &       3332 &          2733 &                             5.22 \\
WIKTIONARY &       3082 &           768 &                            22.57 \\
   ICPCGER &        716 &           715 &                             0.9 \\
 LNC-DE-AT &        188 &           186 &                             0 \\
     Total &     218483 &        110121 &                           100.00 \\
\bottomrule
\end{tabular}
\caption[WUMLS Concept Counts]{Breakdown of proportion of names, concepts and share of matched terms for all source vocabularies present in WUMLS.}
\label{tab:wumls_ontologies}
\end{table}

\textbf{Adapting TLC-UMLS to the German UMLS subset}: When using UMLS\textsubscript{DE} as knowledge base, it is important to remove samples from TLC-UMLS that were annotated with concepts not contained in the German ontologies within UMLS. Keeping these samples would prevent a medical concept normalization system using UMLS\textsubscript{DE} from reaching a theoretical accuracy of 1 on TLC-UMLS. The new subset of TLC-UMLS that contains only concept annotations that are within the German UMLS subset will be called TLC-UMLS\textsubscript{DE}.  It contains 2.988 posts with 6.428 annotations, 200 less than the full TLC-UMLS dataset.

\section{Hardware Setup}

The baseline experiment is run on a 8-core Intel i5-10210U CPU @4.2GHz and 16GB of memory. Training and inference of all other experiments is executed on a Tesla V100 SXM2 GPU @1530MHz with 16 GB of memory.

\section{Baseline}
\label{section:baseline}

The baseline follows the method described in the baseline experiment section by \citet{seiffe_witchs_2020}. The method is based on a string similarity measure to find the closest concept from the UMLS knowledge base for each mention. It can be divided into the following two steps: Extend the German subset with lay terms from an online German lexicon, and create and query a knowledge base with mentions. 

\textbf{Extension of UMLS\textsubscript{DE}:} The German version of Wiktionary was utilized as a second resource to enhance the German UMLS subset with lay terms. As of January 2019, it contained 741,260 entries, some of which belonged to biomedical sub-categories. The German Wiktionary dump from November 2019 was downloaded and processed to filter and parse entries relevant to the biomedical domain, including categories like Medicine, Pharmacy, Pharmacology, Anatomy, etc.. For a full list see Section 5.1. in \cite{seiffe_witchs_2020}. Entries containing the regular expression "krank" (sick) were also included. The final biomedical Wiktionary subset comprised 4,468 concepts, 2,155 of which included at least one synonym, resulting in 8,657 different entries in total. Although smaller than UMLS, Wiktionary offered a valuable range of layman synonyms, such as "Zuckerkrankheit" (sugar disease) or "Zucker" (sugar) for diabetes, which were not listed in UMLS. Other examples include colloquial terms like "Schnelle Katharina" (fast Katharina) and "Flotter Otto" (quick Otto) for diarrhea. To leverage the benefits of both UMLS and Wiktionary, the two datasets were aligned by identifying Wiktionary expressions also present in UMLS. If a Wiktionary term was found within only one UMLS CUI, the term and all its synonyms were added as alternative names for that CUI. If the terms from one Wiktionary entry match multiple UMLS entries, the terms were not included. Ultimately, 768 CUIs were extended with 3,082 additional mentions, resulting in the combined Wiktionary-UMLS (WUMLS) dataset.  

The number of terms added from Wiktionary seems to be relatively low in comparison to the other source vocabularies, i.e.~only $\sim$1.4\% of all terms in WUMLS stem from Wiktionary. Nevertheless, the small portion of Wiktionary terms constitute $\sim$22.57\% of all terms that were matched with mentions. This strongly indicates that the mentions from the TLC dataset deviate from technical medical terms found in UMLS and that Wiktionary is a useful resource to find alternative lay expressions for concepts.  

\textbf{Create and query knowledge base with mentions:} To find the concept with the most similar name for each mention in TLC-UMLS\textsubscript{DE}, a string similarity approach is implemented. Solr (version 8.2), an open-source search engine with fast string similarity search implementations, is used for this purpose. The stemmed WUMLS terms are indexed in a Solr Collection. The similarity measure employed is the edit distance with equal weights for all operations, also known as the Levenshtein Distance.

An example query to find the target concept in the Solr collection for the mention \textit{Pyelonitis} would look like the following: $$\{!func\}strdist("pyelon",index\_term,edit)$$

${!func}$ indicates that a specific search function should be used instead of the default one. The search function keyword comes after the function statement, in this case $strdist$. Next follow the arguments to the search function. The first one is the search term $"pyelon"$, followed by which column to compare the search term to. I named the column that contains the string representation of each concept name $index\_term$. The last argument of \textit{strdist} is the distance function that should be used to compare two strings. The edit distance is specified by $edit$. Other possible distance metrics would be Jaro-Winkler or ngram distance.

\section{Embedding Similarity Methods for MCN}
The prevalent method approach for medical concept normalization primarily involves three key components:

\textbf{Index Embedding Model:} This is the model that produces an embedding representation for all concept names in the knowledge base. In the context of search all embedded concept names are called the index. Notably, each concept can have multiple embedding representations, leading to the situation where the number of index embeddings may exceed the number of concepts, i.e., $|index| \geq |concepts|$. The embedding generation leverages the preferred term of each concept and, if available, its associated synonyms present in the knowledge base.

\textbf{Mention Embedding Model:} This model, similar to the index embedding model, formulates an embedding representation for all mentions. The creation of this embedding can be based solely on the mention itself or may incorporate the surrounding context. Typically, the mention embedding model and index embedding model are the same, ensuring consistency in representation across concepts and mentions.

\textbf{Similarity Metric:} The final ingredient in this approach is a measure that can assess the degree of similarity between each mention embedding and the embeddings in the index. This metric provides the basis for mention-to-concept mapping, which is a central part of MCN.

In this work, I have chosen to experiment with the Transformer-based model, SapBERT\textsubscript{XMLR}, as detailed in \citep{liu_learning_2021}. The choice is justified by the fact that the first version of SapBERT, introduced in this model, is currently recognized as the state-of-the-art in handling English MCN across all primary datasets \citep{liu_self-alignment_2021}. Building upon their monolingual case study the authors have managed to present a multilingual extension of their \textit{Self-Alignment Pretraining} technique \citep{liu_learning_2021}. The general methodology of \textit{Self-Aligment Pretraining} and the application of it to multilingual data can be found in Section \ref{section:self_alginment_pretraining}. 

\citet{liu_learning_2021} report that the model that performed best on a German dataset is the fine-tuned \texttt{bert-german-uncased} model. Nevertheless, I will use SapBERT\textsubscript{XMLR} which lacked behind the German model by 3.7 absolute points in Accuracy@1. Using the German model would compromise the validity of my method to be applied to other languages than German, because for other languages there might be no available base models that had access to a comparable extend of training data in a specific language other than German or English. 

\subsection{SapBERT\textsubscript{XMLR}}
\label{section: sapbert_experiment}
My first experiment shows how the multilingual SapBERT\textsubscript{XMLR} model works in embedding similarity approach on the TLC-UMLS dataset. The model is constructed upon the RoBERTa\textsubscript{XMR} model by \citet{conneau_unsupervised_2020} (Section \ref{section:roberta}). The RoBERTa\textsubscript{XMR} is further refined with the Self-Alignment Pre-training described in Section \ref{section:self_alginment_pretraining}. As knowledge base to be used for the refinement method, the UMLS 2020AB version is used. 
The SapBERT\textsubscript{XMLR} model will be used as index and mention embedding model.
Let $\mathcal{N}$ and $\mathcal{C}$ be all names and concepts in UMLS\textsubscript{DE}. Let $(n,c) \in \mathcal{N} \times \mathcal{C} $ denote a tuple of a concept string representation and its unique identifier, e.g. (\textit{Influenza}, C0021400). Then creating the index is defined as $f: \mathcal{N} \to \Re^{_d}$ where the dimension of the output embedding is $d=1024$ and $f$ is modeled by SapBERT\textsubscript{XMLR} where the \texttt{[CLS]} token is used as output representation. The resulting index matrix is of shape $M \times d$.  

Implementing the formal definition of creating the index, results in the following steps: The \texttt{SapBERT-UMLS-2020AB-all-lang-from-XLMR-large} versions of the Byte-Pair Encoding \citep{sennrich_neural_2016} tokenizer and RoBERTa \citep{phan_robust_2019} model are downloaded from the \href{https://huggingface.co/}{Hugginface Hub} using the \\\texttt{transformers} library \citep{wolf_transformers_2020}. 

\begin{figure}[h]
  \centering
  \includegraphics[width=\linewidth]{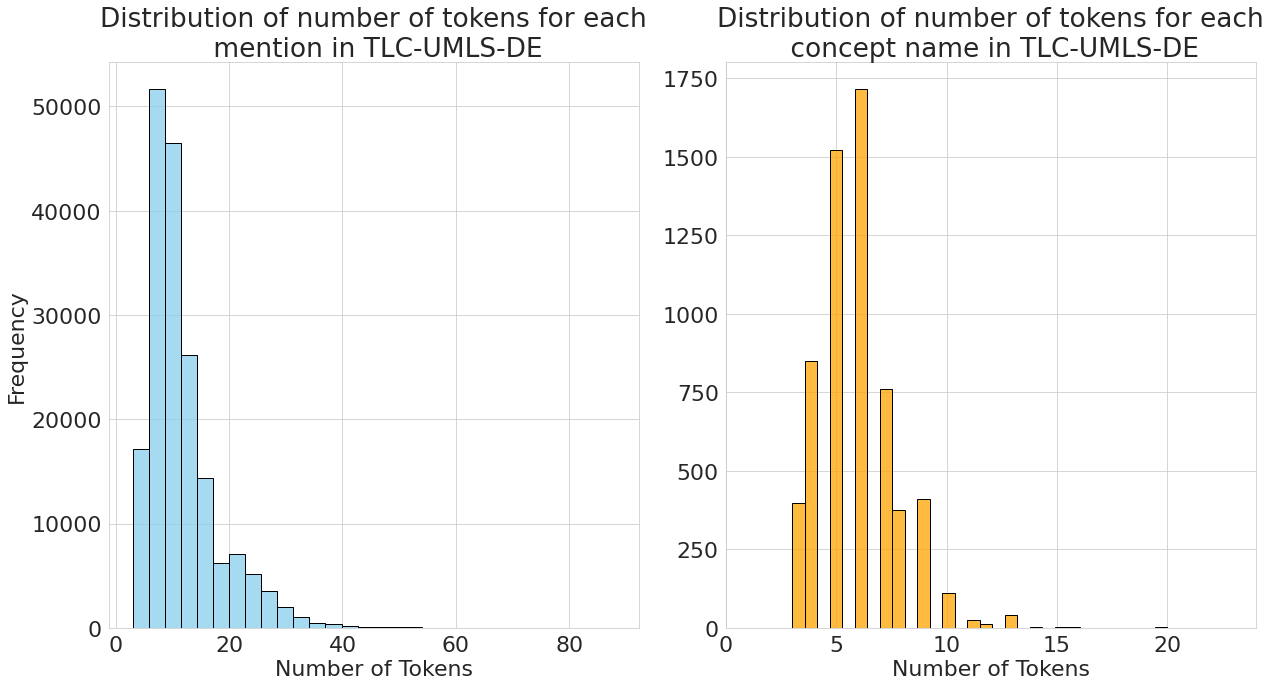}
  \caption[Token Distribution of UMLS\textsubscript{DE} concepts]{The distribution of number of tokens generated by tokenizing concept names from UMLS\textsubscript{DE} and mentions from TLC-UMLS\textsubscript{DE}. As tokenizer the \texttt{SapBERT-UMLS-2020AB-all-lang-from-XLMR-large} tokenizer is used.}
\label{fig:token_lengths}
\end{figure}

To create the index embeddings, the following parameters are used: \texttt{batch\_size=128}, \texttt{padding="max\_length"}, \texttt{truncation=True} and \texttt{max\_length=60}. The maximum output length of the tokenizer is determined by maximum number of tokens that are used to represent on concept name. As one can see from Figure \ref{fig:token_lengths}, most of the concept names are tokenized into a length of one to four tokens, with the majority being around one or two tokens. This suggests that most concept names in the dataset are composed of short phrases or single words. The distribution tapers off for longer token lengths up to a maximum length of 60. By setting the \texttt{max\_length} parameter to the maximum token length in the data, computational resources during inference can be saved because Transformer-based models scale quadratic with the length of the input sequence \citep{vaswani_attention_2017}. The \texttt{batch\_size} is set such that memory usage is maximized, affecting only the inference time.

The tokenizer is applied to all concept names in UMLS\textsubscript{DE} and returns for each name a triple of \texttt{token\_ids}, \texttt{attention\_mask} and \texttt{token\_type\_ids}. The \texttt{token\_ids} are the indices of input sequence tokens in the vocabulary. The \texttt{attention\_mask} indicates for each token if it is a padding token or not so that the attention is only calculated on non-padding token. The \texttt{token\_type\_ids} indicate to which input sequence a token belongs. Since I am working only with single input sequences here, \texttt{input\_type\_ids} will all be set to zero.
The output of the tokenizer is passed to the SapBERT\textsubscript{XLMR} model, which creates the index embedding matrix of shape $215.401 \times 1024$. The computation took $\sim$13 minutes.

To create the mention embedding matrix, the same procedure is applied. The only difference is that \texttt{max\_length} is set to \texttt{20}. The distribution of number of tokens for each mention can also be seen in Figure \ref{fig:token_lengths} where the longest sequence is of length 20. The resulting embedding matrix is of shape $6.428 \times 1024$. The computation took $\sim$16 seconds.

After the creation of index and mention embeddings, the next step is identifying the closest concepts for each mention. I decided to keep the 64 concepts that are closest to a given mention in the index embedding space, so that I can apply the Accuracy@k metric up to \texttt{k=64}. In that way, the error analysis in Section \ref{section:error_analysis} can reason about 64 concepts that were the closest to the mention and find if the correct concept is in them.

To calculate the cosine distance between all mentions and concept names I used the \texttt{spatial.distance.cdist} function from the scipy package \citep{virtanen_scipy_2020}. A similarity function $g$ that uses the similarity function from Section \ref{section:cosine_similarity} can be defined by $$g: f(\mathcal{M}), f(\mathcal{N}) \to \Re^{\mathcal{M} \times \mathcal{N}} $$ where $\mathcal{M}$ is the set of mentions and $\mathcal{N}$ is the set of all names in the knowledge base. I used \texttt{numpy.argsort} to find the indices of the 64 names with the closest distance to each mention $m$. Mapping the indices of the names to the concepts in the knowledge base, yields the respective CUIs for each mention $m$.

\subsection{Choosing the Right Embedding Representation}
\label{section:extraction_configurations}
In the previous experiment, SapBERT\textsubscript{XLMR} was used as embedding model and only the \texttt{[CLS]} token representation was further used. Generally, the model produces a $d=1024$-dimensional representation of each token in the input sequence. The tokenizer adds a special \texttt{[CLS]} token at the beginning of the input and also appends \texttt{[PAD]} tokens to end of the input up to length of 512. The pre-training objective of SapBERT\textsubscript{XLMR}, which is essentially the training of RoBERTa, is designed such that the model learns a \texttt{[CLS]} token representation that contains information about the entire input sequence. The \texttt{[CLS]} token representation can be use to classify the input sequence \citep{phan_robust_2019}. 

During the fine-tuning phase of SapBERT\textsubscript{XLMR} the \textit{Self-Alignment Pretraining} is applied. The performance of Transformer-based and other neural network models tend to decrease when they are fine-tuned with a new task or with new data. \cite{french_catastrophic_1999} first describes the phenomenon known as \textit{catastrophic forgetting}. The work of \citet{kirkpatrick_overcoming_2017} attributes it to the following two causes:
\begin{itemize}
    \item Parameter Overwriting: During fine-tuning, substantial changes in the model's parameters can overwrite the useful features learned during initial training.
    \item Data Discrepancy: When new training data significantly differ from initial training data, the model might change its parameters drastically to accommodate the new data, leading to a loss of previously learned information.
\end{itemize}

In the training phase of SapBERT\textsubscript{XLMR}, both causes of \textit{catastrophic forgetting} are relevant. The embeddings are fine-tuned on the UMLS dataset, which significantly differs from the initial pre-training dataset, the Common Crawl dataset. This dataset is a multilingual and diverse dataset that includes petabytes of scraped web pages, ranging from academic papers and social media texts to coding examples \citep{wenzek_ccnet_2020}. And so it does contain some medical terminology but the vast majority of texts are non-medical.
Simultaneously, the self-alignment pre-training technique only considers the value of the \texttt{[CLS]} token representation, ignoring all other output token representations. Specifically, the multi-similarity loss is calculated using the \texttt{[CLS]} token representation, allowing for a mitigation of any potential loss in the usefulness of this token during fine-tuning. However, despite the focus of the method on the classification token, it is reasonable to experiment with alternative representations.

\citet{vulic_probing_2020} suggest alternative strategies for extraction configurations. They present evidence that, while the \texttt{[CLS]} token can be beneficial for sentence-pair classification tasks, sentence representations derived from \texttt{[CLS]} may be inferior to those achieved by averaging over a sentence’s subwords. In addition to the \texttt{CLS} configuration from Section \ref{section: sapbert_experiment}, they propose two extraction configurations:
\begin{enumerate}
\item \texttt{nospec}: Excludes the special \texttt{[CLS]} and \texttt{[SEP]} tokens from subword embedding averaging.
\item \texttt{all}: Includes both \texttt{[CLS]} and \texttt{[SEP]} tokens in subword embedding averaging.
\end{enumerate}
In order to investigate the impact of these extraction configurations on the performance of SapBERT\textsubscript{XLMR}, the \texttt{nospec} and \texttt{all} configurations will be applied in the same setting as in Section \ref{section: sapbert_experiment}, replacing the use of \texttt{[CLS]} token as the output representation. The execution times remained similar because the dimensionality of all embeddings representation did not change.

It is important to note that an exploration of mixed export configurations, i.e., the cross-product of \texttt{[CLS]}, \texttt{nospec}, and \texttt{all}, as either index or mention embeddings, was also conducted. However, any deviations from homogeneous configurations consistently resulted in a decline in performance.

\subsection{Translating TLC-UMLS to English}
A further experiment I carried out was translating the German TLC-UMLS dataset to English. The authors of of the Self-Alignment Pre-training method \citep{liu_self-alignment_2021} provide an English model to which they applied their technique with the UMLS 2020AB knowledge base. As base model they used PubMedBERT by \citet{gu_domain-specific_2021}. It is noteworthy here, that PubMedBERT is trained using the large version of BERT (Section \ref{section:BERT}). I will refer to this model as  SapBERT\textsubscript{ENG}. 

According to the results published by \citet{liu_learning_2021}, SapBERT\textsubscript{ENG} outperforms SapBERT\textsubscript{XLMR} by one absolute accuracy point on English data. Nevertheless, the difference is rather small and multilingual models should be favored, I want to experiment how well the English model performs on the translated TLC-UMLS dataset.
The create the translated dataset, TLC-ENG\textsubscript{UMLS}, each post in the dataset was translated using the DeepL API\footnote{https://www.deepl.com/docs-api}. To retain the correct start and end of the mention in each translated post, special markers were added before and after each mention in the source data. After translation, the position of the mention between the markers is set as new start and end of the mention. Then the markers are removed.

The experiment setup resembles exactly the setup used in Section \ref{section: sapbert_experiment}. SapBERT\textsubscript{ENG} was used to create the index and mention embeddings and then the closest concepts were searched using cosine similarity.

\section{Does Contextual Information Help?}
\label{section:contex_information}
Contextual information in text is widely recognized to play a crucial role in understanding the semantics of the contained concepts \citep{kim_context-aware_2021, luo_multi-task_2018}. This is particularly important in the field of medical concept normalization where a single term can have multiple meanings depending on the surrounding context, e.g. \textit{cold} can link to a common cold (disease), a temperature or Cold Obstructive Airway Disease. However, the exact impact of context information on the performance of a medical concept normalization model, especially multilingual ones such as SapBERT\textsubscript{XLMR}, remains unclear. \citet{zhang_knowledge-rich_2022} is the only work I found that experiments with SapBERT\textsubscript{ENG} and context information. They embedded a input mention in a mention-centered windows of 64 tokens. Comparing their method to SapBERT\textsubscript{ENG}, they report a accuracy gain of 11 absolute points averaged across five different MCN dataset. Unfortunately, they use different pre-training parameters than the SapBERT\textsubscript{ENG} model, so that they accuracy gain can not solely be attributed to the context information. Moreover, they report an accuracy for the SapBERT\textsubscript{ENG} model that is approximately 20 absolute accuracy points lower than reported in the original paper. This leaves some uncertainty if and to what extend context information really helps in normalizing mentions to UMLS.  
Another reason that justifies conducting experiments with context information is that there is no work present that combines a lay dataset, like TLC-UMLS, with contextual information. This section explores the importance of context in this setting.

\subsection{Embedding Mentions with Context}
In the first of two experiment related to contextual information, not only mention $m$ will be used as input, but $ctx_a, ctx_b$ will added around the mention. There are two assignments that I will try out:

\textbf{Context Window:} Assign the context words around mention $m$ to $ctx_a, ctx_b$, such that the token representation of $(ctx_a, m, ctx_b)$ is of length 64. This is achieved by setting 
\begin{align*}
    len(ctx_a) = \lceil \frac{64 - len(m)}{2} \rceil \\  len(ctx_b) = \lfloor \frac{64 - len(m)}{2} \rfloor
\end{align*}
     
where $\lceil \ \rceil$ is the \textit{ceiling function} and $\lfloor \ \rfloor$ is the \textit{floor function}. The token length of 64 is adapted from \citet{zhang_knowledge-rich_2022}.

\textbf{Sentence Context:} In this setting, the total length of $(ctx_a, m, ctx_b)$ is not fixed, but as context the entire sentence in that $m$ appears is selected. For that spaCy's \citep{honnibal_spacy_2017} rule-based senticizer for German is used. The entire post that $m$ appears is in split into sentences and the sentence that contains $m$ is kept. Then $ctx_a, ctx_b$ are assigned to all the words that appear before and after $m$ in that sentence. Additionally, the start and stop token indices of $m$ are tracked.

The experiment settings are again adapted from Section \ref{section: sapbert_experiment}, but using the context window and sentence context configuration. For the context window variant the \texttt{max\_length} is set to 64 and for the sentence context variant it is set to 150.
I experiment with using the output representation that corresponds to only the tokens of the mention or using the \texttt{[CLS]} token output representation, but the results were consistently worse than averaging over all token output representations. Therefore, these experiments are excluded here.

\subsection{Re-ranking with Contextual Information}
\label{section:cross_encode_training}

Re-ranking is a method employed to refine the initial results produced by a medical concept normalization system \citep{xu_generate-and-rank_2020}. In this process, a set of candidates generated by the system is re-evaluated and sorted. This approach can enhance the model's precision by considering more information than just direct mention-concept similarity, and thus leads to more accurate mappings of mentions to a knowledge base. While the latest and best performing approaches to MCN were embedding similarity-based with no refinement \citep{liu_self-alignment_2021, sung_biomedical_2020, liu_learning_2021}, there exists some interesting re-ranking approaches that are not far behind state-of-the-art \citep{li_cnn-based_2017, wiatrak_proxy-based_2022, mrini_detection_2022, ji_bert-based_2020, mondal_medical_2019, liu_learning_2021}.
\\[12pt]
As alternative to the experiment in section \ref{section:contex_information}, I want to present how contextual information can be used to refine concept candidates that were generated without it. This approach has the advantage that it is highly modular and can be added as a complementary process to most existing medical concept normalization systems. 
As re-ranking model I want to use a sentence cross-encoder model as it is described in \citet{reimers_sentence-bert_2019}. Because there is no multilingual model available that is trained to re-rank medical mentions with context in relation to concept candidates, I will describe how such a model can be trained. 

Firstly, an additional dataset distinct from TLC-UMLS is necessary. Given the lack of other lay German datasets with medical mentions annotated with UMLS concepts, the COMETA dataset \citep{basaldella_cometa_2020} will be utilized. It consists of English medical lay terminology annotated with UMLS, resembling the TLC-UMLS dataset closest in structure.

The COMETA dataset has been divided into 80\% training data and 20\% validation data. There is no test set requirement, as the evaluation will be conducted on the TLC-UMLS data. The cross-encoder should be applied to full sentences from TLC-UMLS that contain a mention. Since COMETA contains already single sentences, there is no need to extract sentences. But to match with the TLC-UMLS dataset, all COMETA training sentences exceeding 150 tokens have been excluded. For each sentence  a set of concept candidates is needed. It consists of the gold CUI for that sentence and 63 other CUIs randomly selected from UMLS\textsubscript{DE}. This results in a training set with 15.900 samples and a validation set of 3.998 samples.

The architecture of the sentence cross-encoder is explained in \ref{section:sentence_cross-encoder}. The multilingual pre-trained language models employed are
 mBERT\citep{devlin_bert_2019} and RobERTa\textsubscript{XMR} \citep{liu_roberta_2019}, together with their respective tokenizers. They are accessible on the Huggingface Hub as \texttt{bert-base-multilingual-cased} and \\ \texttt{xlm-roberta-base}, respectively.

\begin{figure}[h]
  \centering
  \includegraphics[width=\linewidth]{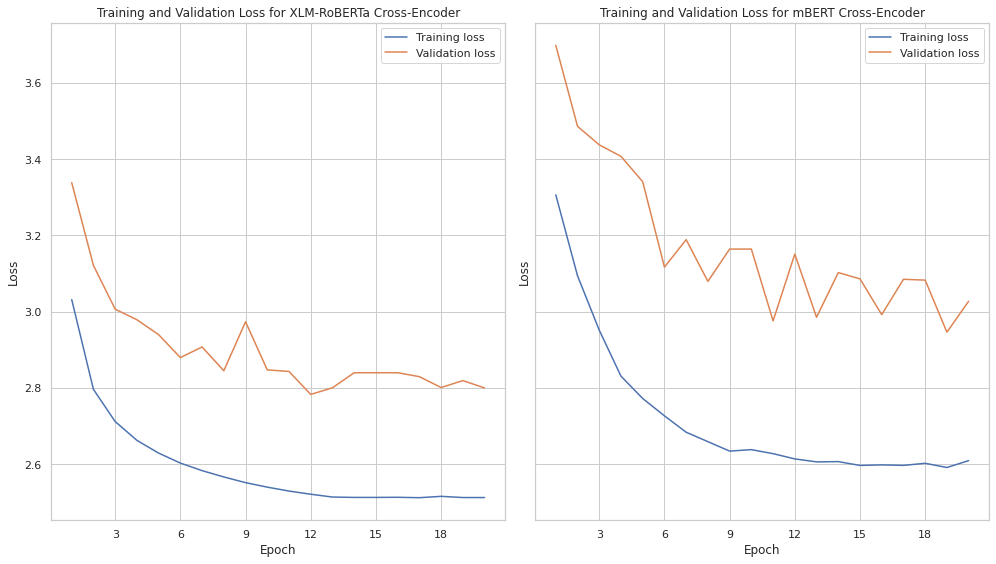}
  \caption[Loss of Cross-Encoders during Training]{Training and validation loss of the RoBERTa\textsubscript{XLM} and mBERT models during training on the COMETA training set. The plots demonstrate how the loss value changes across 20 epochs for both models.}
\label{fig:ce_loss}
\end{figure}

In terms of the training setup, the recommendations of \citet{reimers_sentence-bert_2019} are adopted. These include a batch size of 16, an Adam optimizer with a learning rate of 2e\textsuperscript{-5}, and a linear learning rate warm-up over 10\% of the training data. The pooling strategy is implemented as mean pooling. The model is training for 20 epochs and after each training epoch, the model achieving the best Acc$@$1 on the validation set is saved.
\\[12pt]
The training of each model took approximately 18 hours and 20 minutes.
Figure \ref{fig:ce_loss} presents the learning progress of RoBERTa\textsubscript{XLM} and mBERT Cross-Encoder over 20 epochs. For both models, training loss (blue line) consistently reduces, indicating continuous learning. However, validation loss (orange line) for RoBERTa\textsubscript{XLM} and mBERT Cross-Encoders shows fluctuations from around the sixth epoch onward, reflecting prediction inconsistencies on the unseen validation set. Therefore, while the models improve with training data, their performance on new data varies after about six iterations. 
One can see that the loss curves at epoch one are lower in the left plot. This indicates that the Cross-Encoder with RoBERTa\textsubscript{XLM} as base model performs initially better. This trend continues when the training loss of the RoBERTa\textsubscript{XLM} model decreases monotonically to around 2.55. The training loss on the mBERT model shows some minor fluctuation while it decreases to around 2.6. The RoBERTa\textsubscript{XLM} also consistently over all epochs reaches a lower validation loss than then mBERT model. The lowest validation loss it achieves is $\sim$2.78 while the lowest validation loss of the mBERT model is $\sim$2.96. It is also noteworthy that the validation loss of the mBERT model shows stronger fluctuations indicating that might overfit to the training data. In general, there is always a gap of 0.4 of between the training and validation loss of both models. Since it is consistent from epoch one on, this is a strong indicator that there is a structural difference between the training and the validation set. 

For each training run, the state of the model with the lowest loss on the validation is saved for evaluation. For the RoBERTa\textsubscript{XLM} Cross-Encoder this was at epoch 12 and for the mBERT Cross-Encoder this was at epoch 18. The best RoBERTa\textsubscript{XLM} model reached an Accuracy@1 of 0.71 on the training set and 0.62 on the validation set. The best mBERT model reached an Accuracy@1 of 0.62 on the training set and 0.49 on the validation set.

\chapter{Results}
\label{section:results}

In this section, the examination of the performance of the different models on the TLC-UMLS\textsubscript{DE} dataset is conducted. First, the classification results are presented, followed by a detailed analysis of the re-ranking performance. An in-depth error analysis is provided, identifying common categories of prediction errors.

\begin{table}
\hskip1.0cm\begin{tabular}{|lrrr|}
\toprule
{Model} &  Acc@1 &   Acc@10 & F\textsubscript{1}-score \\
\midrule
Solr + WUMLS                                                            &0.49  &  0.59 & 0.41 \\
\midrule
SapBERT\textsubscript{ENG}\textsuperscript{\dag}                        &0.36  &0.55 & 0.36\\
\midrule
SapBERT\textsubscript{XMLR}\textsuperscript{\dag}                       &0.46  &\textbf{0.78} & 0.44 \\
SapBERT\textsubscript{XMLR}\textsuperscript{\ddag}                      &\textbf{0.51} &0.71 & \textbf{0.50}\\
SapBERT\textsubscript{XMLR}*                                            &0.45 &0.71 & 0.44\\
\midrule
SapBERT\textsubscript{XMLR}\textsuperscript{\dag}  (sentence)           &0.09  &0.25 & 0.12 \\
SapBER\textsubscript{XMLR}\textsuperscript{\dag}  (context window)      &0.01  &0.19 & 0.03\\
\midrule
SapBERT\textsubscript{XMLR}\textsuperscript{\dag} + RoBERTa Re-ranking   &0.21  &0.46 & 0.22\\
SapBERT\textsubscript{XMLR}\textsuperscript{\dag}  + mBERT Re-ranking    &0.09  &0.36 & 0.13\\
\bottomrule
\end{tabular}
\caption[SapBERT\textsubscript{XMLR} Results on TLC-UMLS]{Results of the SapBERT\textsubscript{XMLR} 
models. The \dag, \ddag \ and * mark the \texttt{CLS}, \texttt{all} and \texttt{nospec} extraction configuration from section \ref{section:extraction_configurations}, respectively. }
\label{tab:results}
\end{table}

\begin{figure}
  \centering
  \includegraphics[width=0.8\linewidth]{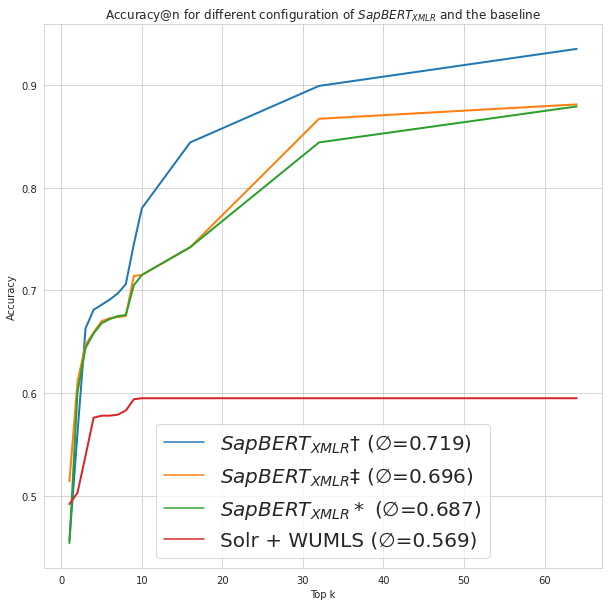}
  \caption[Accuracy@n on TLC-UMLS]{Accuracy@n for different extraction configurations of SapBERT\textsubscript{XMLR} compared to the baseline Solr + WUMLS. The legend also contains the average of Accuracy@n for $n \in [1,...,64]$. For simplicity, only the most interesting models are included in this graph.}
\label{fig:top64_results}
\end{figure}

\section{Classification Results on TLC-UMLS\textsubscript{DE}}

The result are all listed in Table \ref{tab:results}. Starting with baseline model, Solr + WUMLS, which leverages string similarity, exhibits an Accuracy@1 of 0.49 and an Accuracy@10 of 0.59, with an F\textsubscript{1}-score of 0.41. One could say that almost half of the mentions are normalized to the correct concept. Interestingly, the Accuracy@n does not increase for the model for $n$ larger than 9, rather it stays at 0.59 up to Accuracy@64. This can be seen in Figure \ref{fig:top64_results}. The plateau indicates that the baseline method is not able to find the correct concept for a mention even if the allowed maximum edit distance is increased.

The SapBERT\textsubscript{ENG} model on the TLC-UMLS\textsubscript{DE} dataset translated to English reached an Accuracy@1 and F\textsubscript{1}-score of 0.36 and Accuracy@10 of 0.55.

\subsection{Extraction Configuration Experiment Results}
Next the results of the SapBERT\textsubscript{XMLR} model in combination with different extraction configuration is presented. The experiment show what part of the model output works best for representing the mentions and concept names from the knowledge base.
SapBERT\textsubscript{XMLR}\textsuperscript{\dag} that uses the \texttt{CLS} extraction configuration shows an Accuracy@1 of 0.46 and 0.78 for the Accuracy@10. The F\textsubscript{1}-score is 0.36. On the other hand, SapBERT\textsubscript{XMLR}\textsuperscript{\ddag} using the \texttt{all} configuration demonstrates a higher Accuracy@1 of 0.51 and F\textsubscript{1}-score of 0.44, but a lower Accuracy@10 of 0.71. SapBERT\textsubscript{XMLR}\textsuperscript{*}, that makes use of the \texttt{nospec} extraction configuration, is outperformed by the other two models, achieving an Accuracy@1 of 0.45, and Accuracy@10 of 0.71 and a F\textsubscript{1}-score of 0.44. In total, one can observe that the \texttt{all} extraction method outperformed the \texttt{CLS} and \texttt{nospec} extraction method by 0.04 and 0.05 absolute Accuracy@1 points and 0.6 absolute F\textsubscript{1}-score points. 
Figure \ref{fig:top64_results} sheds more light on how close each model is to predict the correct concept. From this figure, it becomes apparent that all three 
 SapBERT\textsubscript{XMLR} models consistently outperform the baseline. One observation that is not apparent in Table \ref{tab:results}, is that the SapBERT\textsubscript{XMLR}\dag \ outperforms the other models consistently on Accuracy@n when at least the top 5 candidates are considered. While the SapBERT\textsubscript{XMLR}\ddag model shows the highest Accuracy@1 and F\textsubscript{1}-score, SapBERT\textsubscript{XMLR}\dag reaches an Accuracy@64 of 0.94. This means that the correct target concept was among the top 64 candidates 94\% of the time. The other two models only reach an Accuracy@64 of 0.88. 

\subsection{Context Experiment Results}

Next, I will present the results of the experiments where not only the mention is used as input but also context of the mention is included. Either in the form of a fixed character window or as the complete sentence the mentions appears. 

When considering sentence and window approaches, the results decrease significantly. SapBERT\textsubscript{XMLR}\textsuperscript{\dag} that uses the complete sentence of the mention to produce the mention embedding has reached an Accuracy@1 of 0.09, Accuracy@10 of 0.25 and a F\textsubscript{1}-score of 0.12. SapBERT\textsubscript{XMLR}\textsuperscript{\dag} (context window) achieves the lowest results with an Accuracy@1 and Accuracy@10 of 0.01 and 0.19. The F\textsubscript{1}-score only is 0.03. In general it is clear that, contrary to the assumption I made in \ref{section:contex_information}, adding context information in the form of surrounding words to the input embedding does not help in normalizing mentions in TLC-UMLS\textsubscript{DE}.

\section{Re-ranking Results on TLC-UMLS\textsubscript{DE}}

Another approach to use context information in the normalization process was proposed in Section \ref{section:cross_encode_training}. The candidate set, produced by the SapBERT\textsubscript{XMLR}\dag model, should be re-ranked by a Sentence Cross-Encoder that gets the full sentence as input.  The same Sentence Cross-Encoder architecture was used but BERT and RoBERTa were used as pre-trained base models.

Both models re-ranked the candidate set and the new ordering produced worse results than the initial candidate set. The RoBERTa Re-ranking exhibits a low performance with an Accuracy@1 of 0.21, an Accuracy@10 of 0.46 and a F\textsubscript{1}-score of 0.22. It still performed better than the mBERT Re-ranking which scored an Accuracy@1 of 0.09, Accuracy@10 of 0.36 and F\textsubscript{1}-score of 0.13. The Re-ranking approach I used shows to only decrease the performance metrics by at least 50\%.

Summing up the results, the embedding similarity approach using the SapBERT\textsubscript{XMLR} model with the \texttt{all} extraction configuration to embed the mention and concept names was able to beat the baseline method by a small margin. Considering only the top 1 predicted concept, the improvement yields only 0.02 Accuracy@1 and 0.09 F\textsubscript{1}-score points. Moreover, it is evident that the embedding similarity approach shows promising results when a set of candidates instead of the top prediction is considered. The improvement over the baseline grows from 0.02 Accuracy@1 points over 0.12 Accuracy@10 points up to 0.28 Accuracy@64 points. The increasing discrepancy shows that using the SapBERT\textsubscript{XMLR} method has the potential to reach much higher performance while the Solr + WUMLS baseline method is capped to a lower performance.

\section{Error Analysis}
\label{section:error_analysis}

Although the best performing method outperforms the baseline method there is still room for improvements. Consequently, I will analyze the top 1 predictions of best performing model, SaPBERT\textsubscript{XMLR} with the \texttt{all} extraction configuration, to shed light on what kind of normalization cases it fails. Initially, a general overview of the mentions that were misclassified will be given, followed by an introduction of five error categories to which these mentions can be assigned.

\begin{figure}[h]
  \centering
  \includegraphics[width=0.8\linewidth]{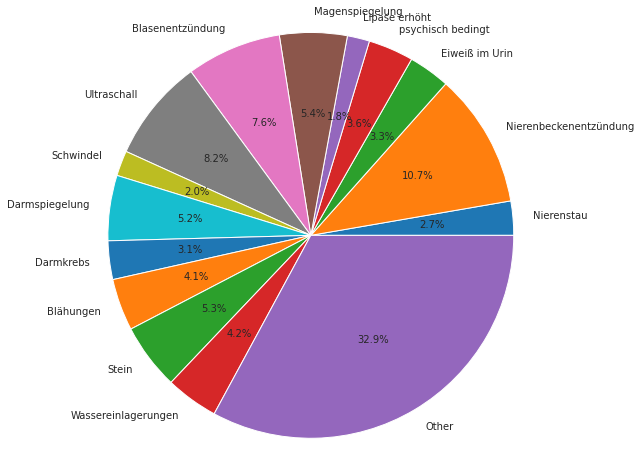}
  \caption[SapBERT\textsubscript{XMLR} Most Common Errors on TLC-UMLS]{All misclassified mentions by SapBERT\textsubscript{XMLR}\ddag \ on TLC-UMLS\textsubscript{DE} with a frequency of at least 50 and their share relative to all misclassified mentions. }
\label{fig:misclassified_mentions}
\end{figure}

The first interesting insight can be gained by looking at the most frequent misclassified mentions. There were a total of 6.228 mentions in the dataset, of which 3.175 were assigned the correct concept and 3.053 assigned the wrong one. Inspecting the most frequent mentions among the misclassified ones reveals that only a few mentions account for a large portion of the errors. By comparing the ten most frequent mentions in the entire dataset, as shown in Figure \ref{fig:most_common_mentions}, to the most frequent misclassified ones, as shown in Figure \ref{fig:misclassified_mentions}, one finds that five out of ten are among them. Namely these were "Ultraschall", "Nierenbeckenentzündung", "Blasenentzündung", "Magenspiegelung", and "Darmspiegelung". Remarkably, these five mentions account for 37.1\% of all errors.

A case worth noting is that ``Steine'' is correctly assigned the concept of \textit{Calculi (C0006756)}, but ``Stein'' (singular of ``Steine'') is wrongly assigned the concept of \textit{Kidney Stone (C0022650)}. Although kidney stones are necessarily calculi, the converse does not hold. For instance, in the sentence \textit{``Nierenkolik ohne Stein und ohne Gries?'' (Renal colic without a stone and without gravel?)} from TLC, the mention ``Stein'' can only be annotated with the general concept \textit{Calculi} and not \textit{Kidney Stone}. Even though the topic of the sentence revolves around the kidney, the more specific concept's assignment is incorrect. Since the model does not use context information for the normalization step, the assignment is too specific.

\begin{figure}[ht]
  \centering
  \includegraphics[width=0.6\linewidth]{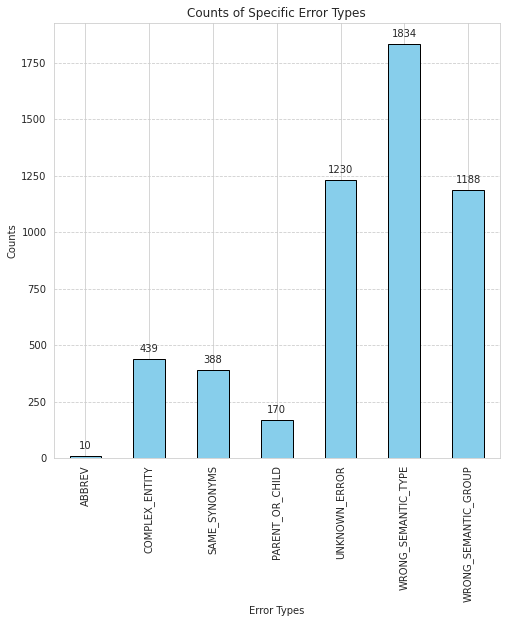}
  \caption[Error Category Count]{Frequency of error types that are assigned to all misclassified mentions by SapBERT\textsubscript{XMLR}\ddag \ on TLC-UMLS\textsubscript{DE}. Note that more than one error type can be assigned to an error.}
\label{fig:error_counts}
\end{figure}

Furthermore, an observation regarding the nature of misclassified mentions is that they are predominantly in lay language. The best performing model, \\ SapBERT\textsubscript{XMLR}\ddag, correctly normalizes 1.870 out of 4.170 lay mentions and 1.430 out of 2.302 technical mentions. This means that only 45\% of lay mentions were correctly classified while 62\% of technical mentions were correctly classified. This underlines the difficulty of the task of normalizing lay mentions to a large knowledge base.

I will further analyze the misclassified mentions by the following 7 error categories\footnote{They are partly inspired by the error analysis in one use case notebook file in the \href{https://github.com/hpi-dhc/xmen/blob/main/notebooks/BioASQ_DisTEMIST.ipynb}{xmen} Github repository.}:
\begin{itemize}
    \item Abbreviation: Triggered if the mention matches the regular expression pattern of having two or three uppercase letters, indicating an abbreviation.
    \item Complex Entity: Caused by a misclassified mention that consists of three or more words, indicating a complex entity that might lead to incorrect mapping.
    \item Same Synonyms: Represents the presence of common synonyms between the predicted concept and the gold standard concept, highlighting an error related to synonymy.
    \item Parent or child concept: Checks if the predicted concept is a parent or child of the gold concept within the UMLS hierarchy.
    \item Unknown Error: Used as a fallback error category when no other specific error type is identified.
    \item Wrong Semantic Type: Triggered if there's no intersection between the semantic types of the predicted and gold concepts, indicating a mismatch in the semantic categorization.
    \item Wrong Semantic Group: Triggered if there's no intersection between the semantic groups of the predicted and gold concepts, indicating a mismatch in the semantic categorization.
\end{itemize}

The resulting distribution of assigning the error categories to all 3.053 misclassified mentions are shown in Figure \ref{fig:error_counts}. Each mention can belong to more than one error category, so the sum of all error categories counts does not add up to 3.053.

The number of misclassified abbreviations were 10. ``KK'' was normalized to (C0202194, \textit{Potassium}) instead of (C0021682, \textit{Health Insurance}). ``KK'' stands for the German word ``Krankenkasse'' which might not be well aligned in the embedding space of SapBERT\textsubscript{XMLR}\ddag, because ``kasse'' is not only the German word for an insurance, as in ``Krankenkasse'', but also the German translation of the more common word checkstand. ``KH'' was also assigned to (C0202194, \textit{Potassium}) instead of (C0019994, \textit{Hospital}). These two errors might be an indicator that the model is biased towards chemical symbols as abbrevations.  

Among the misclassifed mentions were 439 that were categorized as complex entity . The average word count in this category was 3.7. Examples of this category are ``Blutergüsse an der Niere'' classified as (C4049257, \textit{Kidney Congestion}) instead of (C0475022, \textit{Hematoma of kidney}), ``Wasser im Unterleib'' as (C1697454, \textit{Pelvic fluid collection}) instead of (C0003962, \textit{Ascites}), and ``2cm knoten an der schilddrüse'' as (C0749470, \textit{Thyroid cold nodule}) instead of (C0151468, \textit{Thyroid Gland Follicular Adenoma}). All predicted concepts are loosely related to the correct concept, but do not match it exactly. Therefore, the model seems to recognize the individual parts of the mentions, but struggles with assembling them into an complex entity.

There were 388 misclassified examples that belong to the class of same synonyms. These mentions were hard to normalize to the correct concept because there were overlapping synonyms between the correct and predicted concept. The \textit{Self-Alignment Pre-training} technique samples from the synonyms of a concept so that same synonyms bring different concepts closer together instead of separating them. Some examples of the same synonym error category are the following.  ``Harnsäure'' was normalized to (C0041980,\textit{uric acid}) but the correct concept was (C0202239,\textit{uric acid measurement}). ``unfruchtbar'' was assigned the concept (C4074771, \textit{Sterility, Reproductive}), but the correct one was (C0021359, \textit{Infertility}). The mention ``zittern'' was normalized to (C0234369, \textit{Trembling}) while the correct concept was (C0040822, \textit{Tremor}). 

170 out of all misclassified mention were assigned to a concept that is a child or a parent of the correct concept in UMLS. A child concept in the ontology is a concept that is a narrower concept and a parent concept is one that is broader. An example is the mention ``krumme wirbelsäule'' that is normalized to (C0037932, \textit{Curvature of Spine}) which is a parent of the correct concept (C0036439, \textit{Scoliosis}) or ``Hauptschlagader'' that is normalized to (C0003842, \textit{Arteries}) which is a parent of the correct concept (C0003483, \textit{Aorta}). 

The error category that contains the most mentions is the wrong semantic type category, containing 1.834 mentions. It is a superset of the wrong semantic group which contains 1.188 mentions. I will present some examples of the wrong semantic group type but they are simultaneously wrong semantic type error examples. One example is the mention ``Diätologin'' that is normalized to (C0012180, \textit{Dietetics}) that belongs to the semantic type of \textit{Biomedical Occupation or Discipline}. The correct concept would have the semantic type \textit{Professional or Occupational Group}, in this case (C3536818, \textit{Dietitian}). Another one would be the mention ``Ursache'' that is normalized to the concept (C0085978, \textit{Cause}) of semantic type \textit{Idea or Concept} instead of (C0032930, \textit{Precipitating Factors}) of type \textit{Clinical Attribute}.

Lastly, I will present the unknown error category that contains 1.230 misclassified mentions that do not belong to any other error category. To further understand what kind of error exist that are not covered by the previous error categories I inspected the mentions manually. The 1.230 mentions consisted of 168 unique mentions that I assigned to one of five error sub-categories. The counts of each unique mentions are mapped back to how often it occurred in the unknown error sub-category. Three example errors of each of the manual error categories can be found in Table \ref{tab:uknown_categories_example}. I will shortly describe each of the error sub-categories: 

\begin{table}[t]
\hskip-2.0cm\begin{tabular}{@{}|l|l|l|l|l|l|@{}}
\toprule
\makecell{Error\\Category}                     & Example Mention                            & \makecell{Predicted\\CUI} & \makecell{Predicted\\Concept Name}       & CUI      & Concept Name             \\ \midrule
\multirow{3}{*}{\makecell{String\\Similarity}} & Unterleib                          & C0024687      & Unterkiefer                  & C0000726 & Abdomen                  \\
                                   & Nakose                             & C0085349      & Nosema                       & C0278134 & \makecell[l]{Sinnesempfind-\\ungsverlust}     \\
                                   & B12                                & C0062816      & HLA-B13-Antigen              & C0042845 & Cyanocobalamin           \\ \midrule
\multirow{3}{*}{\makecell{Related\\Concept}}   & Blinddarmdurchbruch                & C0341401      & Blinddarmperforation         & C0854119 & \makecell[l]{Rupturierte\\Appendizitis} \\
                                   & Ständiges Wasserlassen             & C0375553      & Ueberlaufinkontinenz         & C0032617 & Polyurie                 \\
                                   & \makecell[l]{hormonproduzierender\\Tumor}         & C0027661      & \makecell[l]{hormonabhängige\\Tumore}     & C0206754 & \makecell[l]{Neuroendokrine\\Tumoren}   \\ \midrule
\multirow{3}{*}{\makecell{Negated\\Mention}}   & \makecell[l]{nicht-bakterielle\\Blasenentzündung} & C0742964      & bakterielle Zystitis         & C0282488 & \makecell[l]{intersistielle\\Zystitis} \\
                                   & \makecell[l]{nicht künstlicher\\Darmausgang}      & C0042020      & \makecell[l]{künstliche\\Harnableitung}     & C0014370 & Enterostomie             \\
                                   & symptomfrei                   & C0586406      & Augensymptom & C0436342 & Keine Symptome       \\ \midrule
\multirow{3}{*}{\makecell{Retired\\CUI}}       & Herzrasen                          & C1868917      & Herzrasen                    & C9938231 & Tachykardie              \\
                                   & obstipation                        & C0012616      & Exartikulation               & C0009806 & Verstopfung              \\
                                   & \makecell[l]{Fruchtwasser-\\untersuchung}           & C0013690      & Hydrarthrose                 & C0002627 & Amniozentese             \\ \midrule
\multirow{3}{*}{\makecell{Unrelated\\Concept}}       & Abspecken                          & C0444186      & Abstrichuntersuchung                    & C1262477 & Gewichtsverlust              \\
                                   & Pobacke                        & C1504061      & Citrullus colocynthis               & C0006497 & Gesäß              \\
                                   & \makecell[l]{Herzstolpern}           & C0996952      & Leonurus                 & C0033036 & \makecell[l]{Atriale\\Extrasystoliekomplexe}             \\ 
                                   \bottomrule
\end{tabular}
\caption[Error Category Examples]{The table shows three examples of each error sub-category of the unknown error category. The concept names are in German and show the concept names that are actually used to build the search index.\label{tab:uknown_categories_example}}
\end{table}

\begin{description}
    \item[String Similarity:] Mentions in this sub-category are assigned to a concept that has a high string similarity, meaning parts of the mention and the predicted concept are similar. There were 109 such cases in total. The predicted concept of the mention ``Unterleib'' is (C0024687, \textit{Unterkiefer}). The model likely confused the similarity between the sub-strings ``Unter'' and ``Unter'' in the two terms, leading to this error.
    
    \item[Related Concepts:] This sub-category contains mentions that are related to the predicted concept but not identical. 776 mentions belong to this sub-category. These errors highlight the difficulty of MCN in distinguishing nuances between closely related medical concepts. ``Blinddarmdurchbruch'' is normalized to (C0341401 , \textit{Blinddarmperforation}) instead of (C0854119, \textit{Ruptuierte Appendizitis}). Both terms are related to complications of the appendix, but the model struggled to differentiate their specific meanings.

    \item[Negated Mention Errors:] 21 mentions fall into this sub-category. These errors stem from negations, where the model struggles to comprehend the negation and assigns a concept opposite to the intended meaning. The mention ``nicht-bakterielle Blasenentzündung'' was assigned the concept (C0742964, \textit{bakterielle Zystitis}). The negation ``nicht'' in the mentions was not correctly understood by the model and led to a wrong prediction.

    \item[Retired CUI Errors:] This sub-category comprises concepts that have been retired from the UMLS knowledge base but still appear in the training data. Approximately 21 mentions belong to this sub-category. Mention ``obstipation'' is normalized as (C0012616, \textit{Exartikulation}). The retired concept C0012616 does not exist in the 2022AB version of UMLS anymore, therefore it can not be predicted by the model that uses the index with concept names from the 2022AB version. 

    \item[Unrelated Errors:] Finally, there were 300 errors for which I could not find a summarizing error sub-category. There is no clear relation between the predicted and the actual concept and their surface form are vastly different. The mention ``Abspecken'' is classified as (C0444186, \textit{Abstrichuntersuchung}) but not (C1262477, \textit{Gewichtsverlust}). The terms have very different meanings, and the model struggled to find any meaningful connection between them. The mention ``Pobacke'' and ``Herzstolpern'' are classified as (C1504061, \textit{Citrullus colocynthis}) and \\ (C0996952, \textit{Leonurus}). Both predicted concepts are some form of plant genus unrelated to the mentions. 
\end{description}

The introduction of the error sub-categories concludes the error analysis and also the result section. Before discussing the results, I will shortly summarize the key points of this section. 

The results section examines the performance of various models on the TLC-UMLS\textsubscript{DE} dataset. The classification outcomes are presented for different models, including the baseline Solr + WUMLS and the SapBERT\textsubscript{XMLR} model with varied extraction configurations. The baseline method reveals a limitation in finding the correct concept beyond a certain threshold. Among the SapBERT\textsubscript{XMLR} models, the \texttt{all} extraction method outperformed others on Accuracy and F\textsubscript{1}-score, and they consistently surpassed the baseline. Context experiments, contrary to expectations, resulted in a decline in performance. Finally, the re-ranking approach using Sentence Cross-Encoders also produced disappointing outcomes, reducing metrics by at least 50\%. 

In summary, the embedding similarity approach using SapBERT\textsubscript{XMLR} with the \texttt{all} configuration slightly beats the baseline, showing promising results when considering a set of candidates, while other methods either do not add value or decrease performance. In Subsection \ref{section:error_analysis}, the error analysis of SaPBERT\textsubscript{XMLR} with the \texttt{all} extraction configuration is conducted. The section begins with an overview of the misclassified mentions and introduces five main error categories. Insights into the most frequent misclassified mentions reveal that a few mentions account for a significant portion of the errors, and the misclassifications often involve lay language. 

Seven specific error categories are analyzed. Examples for each category are provided, and the analysis uncovers specific trends and challenges, such as the model's struggles with assembling complex entities and bias towards certain interpretations, like chemical symbols as abbreviations. The analysis concludes with a manual inspection of the unknown error category, further dividing it into sub-categories such as string similarity, related concepts, and negated mentions.

\chapter{Discussion}
\label{section:discussion}
In this section I will connect the initial research question of this thesis with the methodology I explained in Chapter \ref{section:experiments} and the results from Chapter \ref{section:results}. Additionally, I will include potential strategies to mitigate identified error categories from the former Section \ref{section:error_analysis}.

\section{Normalizing German Lay Mentions}

In this section, the focus lies on the outcomes of the different experiments. I will put each outcome into context and briefly discuss the findings.

\textbf{The Baseline:} The strong performance of the baseline, despite only utilizing simple string similarity, emphasizes the variety and completeness of concept names and synonyms in WUMLS. One observation from Table \ref{tab:results} is that string-based similarity methods hit a performance limit at almost 0.5 Accuracy. This seem to indicate that half of the mentions in TLC-UMLS do not correspond to a concept name or synonyms in WUMLS with similar surface form. String-based similarity methods fail to normalize these mentions since they require contextual or semantic information. To support this claim empirically, I calculated the edit distance between each correctly normalized mention and the concept name it was normalized to. Additionally, I normalized the values by the length of the longer string of the mention and concept name. This results in a normalized edit distance of 0.12 between the mentions and the correctly normalized concept names. This means that in general 12\% of the longer string needs to be edited to equal the other string, which is not much. Doing the same with the correct predictions of the SapBERT\textsubscript{XMLR}\ddag model results in a normalized edit distance of 0.62.  
This shows the advantage of contextualized approaches like SapBERT, capable of normalizing mentions that are orthographically different than the names of the concept they belong to.

\textbf{Performance of SapBERT as MCN Model:} Modern Transformer-based models, fine-tuned in an unsupervised fashion on a knowledge base, demonstrate significant efficacy in normalizing medical concepts. As portrayed in Figure \ref{fig:top64_results}, SapBERT\textsubscript{XMLR}\ddag\ displays the highest Accuracy@n for $n\leq5$, and for $5<n<64$ SapBERT\textsubscript{XMLR}\dag\ reaches the highest scores. The error analysis is based on the former model because the analysis only considers the top prediction. Nevertheless, the SapBERT\textsubscript{XMLR}\ddag\ model shows the greatest potential for future improvement because it has an Accuracy@64 of 0.93, which is $\sim$0.05 absolute points higher than the one from SapBERT\textsubscript{XMLR}\dag. 

Furthermore, the multilingual models bridge the performance gaps to monolingual models. Interestingly, translating the dataset to English and applying the English SapBERT model yielded results below the baseline. As noted by \citet{liu_learning_2021}, the multilingual SapBERT\textsubscript{XMLR} model lacks only 1 absolute Accuracy point behind monolingual SapBERT\textsubscript{ENG} on English data. 

\textbf{Embedding Context Information for Similarity Search:} The study of \citet{basaldella_cometa_2020} suggest that neural models can exploit context information to resolve difficult normalization samples. Contrary to expectations, the incorporation of context information did not improve the performance of the normalization step in the experiments. The two implemented approaches using the sentence context and context window approach failed to produce satisfying results compared to no context or the baseline. However, extending the context window to entire sentence to a maximum of 150 tokens did result in a minor improvement over shorter context lengths. This suggests that the general methodology on how context information was utilized needs refinement, but using a larger context window than 64 tokens seems to help.

\textbf{Re-Ranking with Sentence Cross-Encoders:} The most similar experiment to my re-ranking experiments were conducted by \citet{lin_enhancing_2022}. They also use a sentence cross-encoder to refine a set of candidates for medical concept normalization. The difference is that they used the same mention representation without context for candidate generation and re-ranking. They report that the re-ranking approach is on average on par with the direct candidate generation approach. It can be noted that the approach was not very promising without context information. Connecting this to my experiment, the results were even worse when context information was added. Both approaches I implemented reached bad results in comparison to no context or the baseline, but extending the context window to a maximum of 150 tokens results in an F\textsubscript{1}-score of 0.12 over 0.03. Conclusively, more context information seems to help but the general approach on how context information was used in this thesis did not work well.

Concluding the discussion, the findings suggest that deep learning models, specifically SapBERT, perform significantly better than traditional string-based similarity methods. While the baseline performance is robust, it highlights the limitations of non-contextual approaches, especially their inability to capture semantic nuances. SapBERT's superior performance underscores the importance of embedding semantic information into the normalization process. However, the integration of context information requires further refinement to optimize its efficacy.

\section{Error Mitigation Ideas}
\label{section:error_mitigation}
The error analysis reveals several insights into the misclassified mentions by \\SapBERT\textsubscript{XMLR}\ddag \ on TLC-UMLS\textsubscript{DE}. From the investigation of the seven defined error categories and the manual inspection of the unknown error category, the following conclusions are drawn:

\textbf{Semantic Understanding:} A significant number of errors can be attributed to the ambiguity of mentions and type mismatches. For example, a term like "cold" could refer to a viral infection or a low temperature. This type of error indicates that the model has difficulty distinguishing between different types of medical concepts. The issue relates to the ambiguity of mentions, and there are approaches in the related work section that use semantic type information to tackle this problem. Specific techniques that leverage semantic type prediction can be employed to enhance the model's understanding of concepts. A semantic type prediction component can be realized as described by  \citet{vashishth_improving_2021}. It re-ranks candidates based on the likeliness of their semantic types in relation to the mention. The system of \citet{zhu_latte_2020} works independently of candidates and predicts a semantic type given a mention. The predicted semantic type could be used to narrow down the set of possible concepts. For example, instead of the full search index of all concept names, a search index of only a subset of concept names that have the predicted semantic types could be used.

\textbf{Fine-Grained Distinctions:} All errors from the related concepts and parent-or-child category indicate a challenge in making fine-grained distinctions between closely related medical concepts. The inherent complexity of medical concepts makes it a hard problem to map mentions, especially lay, to the correct concept. Such errors might be reduced by enhancing the training data with more nuanced examples. Such sampling process could be incorporated in the sample selection process of SapBERT. Furthermore, additional context could help to disambiguate related target concepts better. The context information could be used in the MCN model directly or a specialized disambiguation technique can be applied to the set of concept candidates.
\newline

Rethinking the evaluation metrics could also solve the aforementioned cause of errors. Accuracy@n already loosens the strict hit-or-miss approach of standard Accuracy, but it does not take the state of the knowledge base into account. By leveraging the hierarchical structure of UMLS, one could design an evaluation metric that quantifies the quality of a prediction by the number of edges between the predicted and true concept. It would also be interesting to use this graph-based evaluation metric as a loss metric. Potentially, it could be used as a counterpart to the cosine similarity by operating in the graph-based space of the knowledge base. 

Independently of how a graph-based evaluation would look like, it makes sense to ask the question what degree of precision is sufficient for MCN use cases. As stated in Chapter \ref{section:introduction}, MCN results can be used to gain health-related insights from unstructured text. For example, these can be useful for improving search results but also for identifying patterns such as risk factors or adverse drug reactions. Arguably, the UMLS knowledge base has a much higher resolution of medical concepts than is needed to accomplish these tasks. For example, for investigating certain symptoms of a disease, the presence of a condition might be more important than the exact location. Another example would be finding adverse drug reactions for which it might be more important to reliably detect them, even if the predicted concept is not the most narrow concept that could be assigned.

\chapter{Conclusion and Future Work}
\label{section:conclusion}
This section concludes my thesis by summarizing key 
findings, acknowledging limitations, and proposing directions for future work. It aims to highlight the contributions made towards the normalization of German medical lay terms in user-generated texts.

\section{Conclusion}
The main goal of this thesis is to investigate how current medical concept normalization methods perform in a low resource setting. For that I created dataset that comprises German lay data obtained from an online medical forum and annotated it with UMLS concepts. A simple but effective baseline that can be used to test MCN systems against the dataset was established. Furthermore, I created an overview of different approaches to MCN that exist in the literature. I implemented one recent multilingual state-of-the-art approach from the literature that works with German data. Based on that approach I evaluate whether context information is helpful in lay mentions. Seven distinct errors were created that I used to examine the predictions of the best performing system more in detail. As last contribution, I pointed at methods or improvements that could mitigate a large portion of errors.

TLC-UMLS is the first dataset that provides UMLS annotations for medical mentions in German user-generated texts. It contains 6.428 annotated mentions and more than two thirds of the mentions are lay language. The dataset provides a valuable resource to understand how medical concepts are expressed in non-technical terms.

Recent multilingual MCN methods \citep{yuan_coder_2022, liu_learning_2021} show strong performance across different languages. The \textit{Self-Alignment Pre-training} approach is able to refine the RoBERTa\textsubscript{XMR} model such that its word embeddings can be used to normalize German lay mentions to the Unified Medical Language System. The performance is comparable to other multilingual MCN studies but it is still inferior to approaches that exclusively focus on English models and data.
On top of the SapBERT model, different extraction configuration were tested. Using the mean of the complete token output representation of the model yielded the highest Accuracy@1. Increasing the number of candidates to over five instead of considering only the top results showed that using the \texttt{CLS} token representation as representation of the mentions yielded better results.

Lay mentions come with some unique challenges. The methods I tested were not able to reach results that these methods reached on technical texts. The challenges that come with German lay mentions are explained in Section \ref{section:error_analysis} and validated by the experimental results in Section \ref{section:results}.
One challenge of lay mentions is the vague nature of lay language. To tackle it, I conjectured that context information would help normalization in general, but especially with lay language. This conjecture proved wrong in this case. I tried to create mention embeddings from mentions with context (Section \ref{section:contex_information}). Alternatively, a Sentence Cross-Encoder was trained from scratch to re-rank a set of concept candidates based on the context of a mention. Both methods produces results that were far below the baseline. My conclusion of the context experiments is that both methods do not use the context information in a useful way to improve MCN.

In Section \ref{section:error_mitigation}, I pointed at two common errors that were made by the best performing MCN model. I also laid out how a different evaluation approach could better reflect the actual desired performance of a MCN system. Instead of simply rejecting a wrong concept normalization, the hierarchical structure of UMLS should be leveraged to give a more nuanced feedback on the quality of the normalization.
Moreover, I showed that the embedding similarity approach with multilingual SapBERT offers a suitable solution for normalizing mentions in German UGTs. 

\section{Limitations}
Like any other system or framework, the presented MCN methods have some limitations as well. In the following, I will talk about the limitations of my work:
\begin{enumerate}
    \item The TLC dataset still contains many mentions that are not annotated. In this thesis the pre-annotated mentions were annotated with UMLS concepts. But the pre-annotations were already incomplete. 
    
    \item Evaluating the separation of concepts in the embedding space remains a challenge. Given the vast amount of unique medical concepts, achieving high coverage of concepts in UMLS would necessitate an extensive volume of annotated text data. An alternative for evaluating this separation could be sampling directly from the UMLS to assess the distribution of different concept clusters within the embedding space.

    \item The model has limitations in handling ambiguous mentions, particularly when contextual information is absent. For instance, in the case of the term "cold," the current implementation will default to linking it to (\textit{Common Cold}, C0009443), excluding other plausible mappings like (\textit{Cold Temperature}, C0009264). To resolve such issues, the inclusion of context-aware methods is indispensable.
\end{enumerate}

\section{Future work}

In this section, future research directions aimed at improving medical concept normalization systems are outlined.

\textbf{Create Data Annotation Specifications for MCN:} Developing a comprehensive annotation guideline for medical concept normalization is crucial for ensuring consistency and quality in data labeling. These specifications should detail how to handle various cases, such as ambiguous mentions, mentions with multiple possible concept IDs, or mentions that are part of a larger phrase. Establishing such guidelines can significantly streamline the annotation process and improve the robustness of MCN systems.

\textbf{Adding context information to SapBERT pre-training:} The current SapBERT method samples concept names from the UMLS. One could scrape from medical resources or use a generative NLP model to generate examples sentences that contain mentions in UMLS. Presumably, this method would require more samples for training since the context would also introduce noisy samples. But in that way the refinement method SapBERT would be context-sensitive. The work of \citet{zhang_knowledge-rich_2022} showed how text samples that contain mentions can be samples from scientific papers.

\textbf{Semantic Type Prediction:} To avoid semantic type errors a semantic type prediction component could be added to the normalization step. Either as post-processing of the candidate list, e.g. like the system proposed by \citet{vashishth_improving_2021}. Alternatively, the loss function used for SapBERT can be adapted so that it penalizes wrong semantic type prediction. It can be used in the similarity search without further changes. 

\textbf{More synonyms in UMLS:} Similarly to how the WUMLS dataset was constructed, the list of concept names for a concept can be extended. Wiktionary proved as a valuable resource, especially for synonyms in lay language. The effectiveness of Wiktionary was shown by the fact that 22.57\% of mentions correctly normalized by the baseline were linked to concept names that stem from Wiktionary.
\newline

The outlined directions address the identified shortcomings. Implementing these ideas may significantly enhance the reliability and accuracy of future medical concept normalization systems.

\printbibliography[heading=bibintoc]

\end{document}